\documentclass{article}

\usepackage{arxiv}

\usepackage[utf8]{inputenc} 
\usepackage[T1]{fontenc}    
\usepackage{hyperref}       
\usepackage{url}            
\usepackage{booktabs}       
\usepackage{amsfonts}       
\usepackage{nicefrac}       
\usepackage{microtype}      
\usepackage{graphicx}
\usepackage{natbib}
\usepackage{doi}
\usepackage{cite}
\usepackage{multirow}
\usepackage{subcaption}
\usepackage{longtable}
\usepackage{multicol}
\usepackage{amsmath, xparse}
\usepackage{enumitem}
\newcommand{\tightlist}{%
\setlength{\itemsep}{0pt}\setlength{\parskip}{0pt}}

\title{Computer-aided Interpretable Features for Leaf Image Classification}

\author{ {\hspace{1mm}P. G. Jayani Lakshika} \\
	Department of Statistics\\
	University of Sri Jayewardenepura\\
	Nugegoda, CO10250, Sri Lanka \\
	\texttt{jayanilakshika76@gmail.com} \\
	\And
	{
	\hspace{1mm}Thiyanga S. Talagala} \\
	Department of Statistics\\
	University of Sri Jayewardenepura \\
	Nugegoda, CO10250, Sri Lanka\\
	\texttt{ttalagala@sjp.ac.lk} \\
}

\renewcommand{\shorttitle}{Computer-aided Interpretable Features for Leaf Image Classification}

\hypersetup{
pdftitle={Computer-aided Interpretable Features for Leaf Image Classification},
pdfsubject={q-bio.NC, q-bio.QM},
pdfauthor={P. G. Jayani Lakshika, Thiyanga S. Talagala},
pdfkeywords={Medicinal,  Image processing, Feature extraction, LDA, PCA},
}

\begin{document}
\maketitle

\begin{abstract}
	Plant species identification is time-consuming, costly, and requires lots of effort, and expert knowledge. The studies on medicinal plant identification are often performed based on medicinal plant leaf images. The is because plant leaves contain a large number of diverse sets of features such as shape, veins, edge features, apices, etc. that are useful in identifying different medicinal plants. In recent, many researchers use deep learning methods to classify plants directly using plant images. While deep learning models have achieved great success, the lack of interpretability limits their widespread application. To overcome this, we explore the use of interpretable, measurable, and computer-aided features extracted from plant leaf images.   Image processing is one of the most challenging, and crucial steps in feature extraction. The purpose of image processing is to improve the leaf image by removing undesired distortion. The main image processing steps of our algorithm involve: i) Convert original image to RGB (Red-Green-Blue) image, ii) Gray scaling, iii) Gaussian smoothing, iv) Binary thresholding, v) Remove the stalk, vi) Closing holes, and vii) Resize the image. The next step after image processing is to extract features from plant leaf images. We introduced 52 computationally efficient features to classify plant species. These features are mainly classified into four groups as i) shape-based features, ii) color-based features, iii) texture-based features, and iv) scagnostic features. Length, width, area, texture correlation, monotonicity, and scagnostics are to name a few of them. We explore the ability of features to discriminate the classes of interest under supervised learning and unsupervised learning settings.  For that, supervised dimensionality reduction technique, Linear Discriminant Analysis (LDA), and unsupervised dimensionality reduction technique, Principal Component Analysis (PCA) are used to convert and visualize the images from digital-image space to feature space. All the applications are performed on Flavia and Swedish open-source benchmark leaf datasets. The results show that the features are sufficient to discriminate the classes of interest under both supervised and unsupervised learning settings. The results of this study are beneficial for the researchers working in the field of developing automated plant identification and classification systems.
\end{abstract}


\section{Introduction}

    Leaf identification is becoming very popular in classifying plant species. Plant leaf contains a significant number of features that can help people to identify and classify plant species. From a medical perspective, medicinal plants are usually identified by practitioners based on years of experience through sensory or olfactory senses. The other method of recognizing these plants involves laboratory-based testing, which requires trained skills, data interpretation which is costly and time-intensive. Automatic ways to identify medicinal plants are useful especially for those that are lacking experience in medicinal plant recognition. Statistical machine learning techniques play a crucial role in the development of automatic systems to identify medicinal plants. In developing such a system input features play an important role. The main aim of this paper is to introduce interpretable features that can be computed based on plant leaf images. Image processing and feature extraction play crucial roles in developing a workflow to achieve the aim of this research. 

  The main aim of image processing is to extract important features by removing undesired noise and distortion \citep{articlee}. Image processing steps include image segmentation \citep{DBLP}, image orientation, cropping, grey scaling, binary thresholding, noise removal, contrast stretching, threshold inversion, image normalization, and edge recognition are some of the image processing techniques applied in recent research. These steps can be applied parallel or individually, several times until the quality of the leaf image reaches a specific threshold. 

 The second step is feature extraction, which identifies and encodes relevant features from leaf images \citep{articlee}. This is a challenging task due to the structural diversity of the leaf images. Therefore, in recent years,  many researchers use deep learning methods to classify plants directly using plant images \citep{4458016,articlepl,inproceedings}. While deep learning models have achieved great success, most models remain complex black boxes.  The lack of interpretability limits their widespread application.

   A digital image is a combination of pixels from three different color planes red, green, and blue. Images are stored in computers as three separate matrices corresponding to the red, green, and blue color channels of the image. The three separate matrices corresponding to intensities of colors at different positions of the image \citep{book1}.

  The main aim of feature extraction is to reduce the dimensionality of this information by obtaining measurable patterns of leaf images. For example, shape, color, and texture are some of the patterns that may be observed. In this paper, we introduce a collection of interpretable, measurable, and computer-aided features that are useful in image leaf classification. This feature collection includes several pre-established features identified through a thorough review of the literature.  Other than existing features, we introduce several features computed based on the Cartesian coordinate of the images.  Furthermore, we explore the ability of features to discriminate the class of interest under a supervised learning setting and an unsupervised learning setting.

     The paper is organized as follows. Section 2, describes the steps of image processing. Preprocessing is necessary before extracting features from the images. Section 3, discusses feature extraction in in-detail because features are highly influenced by the plant species to be classified. Under this section, we discuss mainly four types of features as shape, color, texture, and scagnostics and how to extract them. Section 4, presents empirical application. This section consists of details about the datasets that are used to explain the applications, and visualization of leaf images in the feature space using supervised, and unsupervised dimensionality reduction techniques. Section 5, consists of a summary of the software and packages used to extract the features. Some discussion about the outputs and concluding remarks are given in the last section.

\section{Image Processing}

\begin{figure}[!ht]
{\centering \includegraphics[width=0.5\linewidth]{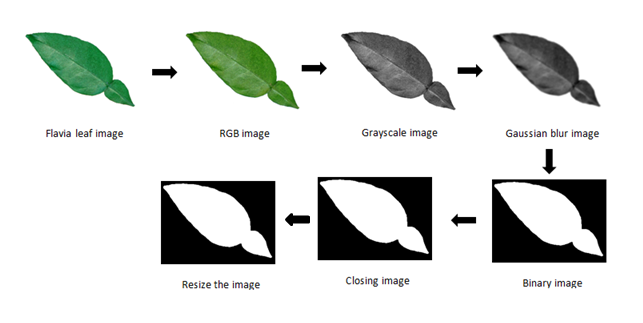} 

}

\caption{\label{fig:test2}Image processing workflow.}\label{fig:unnamed-chunk-2}
\end{figure}

      Image processing is an essential step to reduce noise and content enhancement while keeping its features intact \citep{8675114}. The workflow we use to process images in this paper is shown in Figure \ref{fig:test2}. This includes seven main steps. They are: i) converting BGR (Blue-Green-Red) image to RGB (Red-Green-Blue), ii) gray scaling, iii) Gaussian filtering, iv) binary thresholding, v) remove stalk, vi) close holes, and vii) image resizing. Some of these steps are applicable only to specific images. For example,  apply to remove stalk is applicable only to leaf images that have a stalk.

\begin{figure}[!ht]
\begin{subfigure}{.5\textwidth}
\centering
\includegraphics[width=60mm, height=20mm]{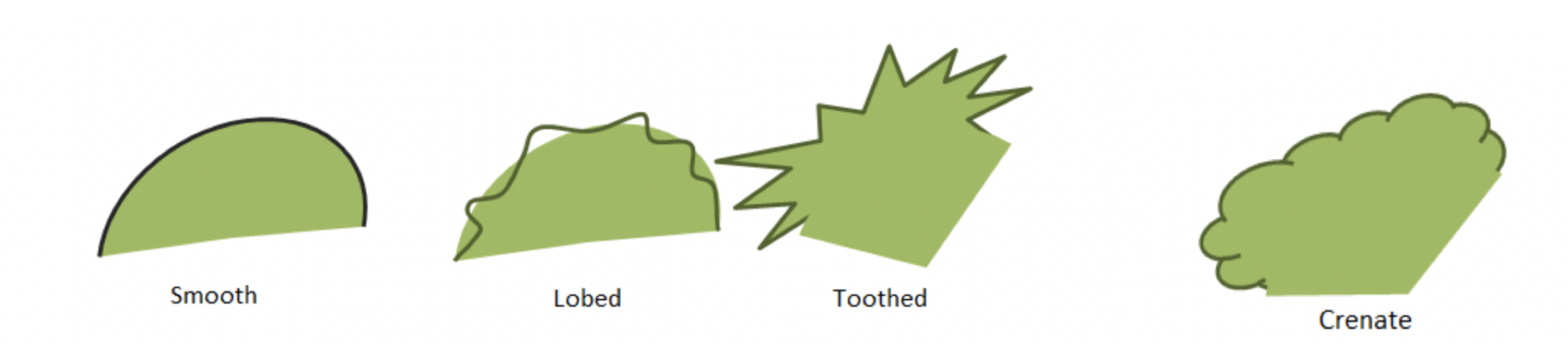}
        \caption{\label{alledge} Edge types of leaves focus in the study.}
\centering
        \includegraphics[width=80mm, height=30mm]{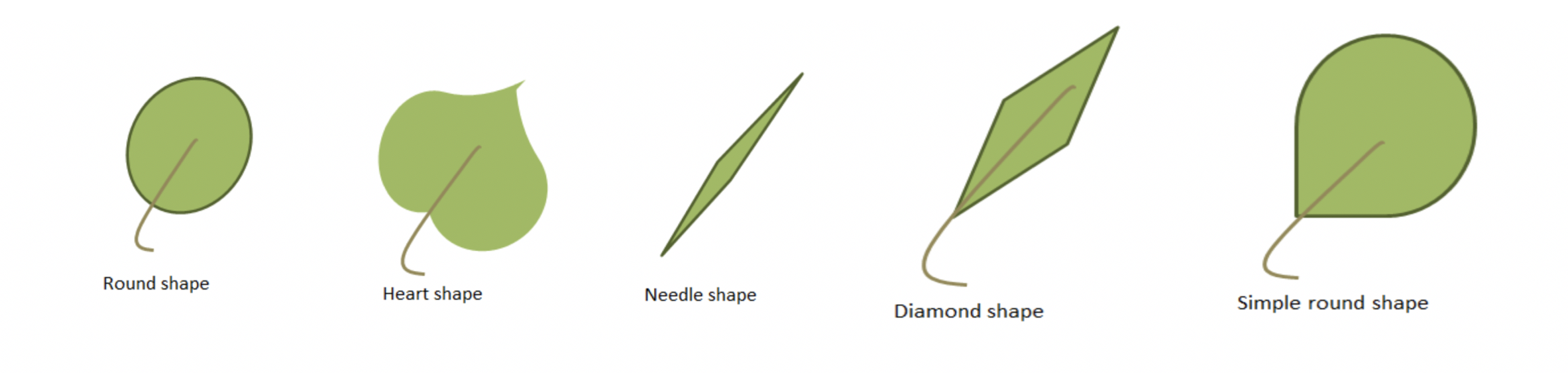}
        \caption{\label{shapeimg}Shape illustrations}       
        
\end{subfigure} 
\begin{subfigure}{.5\textwidth}
\centering
        \includegraphics[width=50mm, height=50mm]{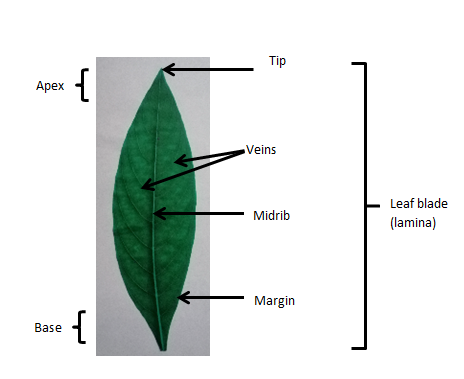}
        \caption{\label{simplearra} A leaf with a simple arrangement.}
        
\end{subfigure} 

\caption{Leaf feature structures considered in the analysis}
        \end{figure}

In this study, we focus on leaves with a simple arrangement as shown
in Figure \ref{simplearra}. A single leaf that is never divided into
smaller leaflet units is known as a leaf with a simple arrangement. This
type of leaf is attached to a twig by its stem or the petiole. The
margins, or edges, of the leaf, can be smooth, lobed, or toothed (see
Figure \ref{alledge}).

\hypertarget{step-1-converting-bgr-blue-green-red-image-to-rgb-red-green-blue}{%
\subsection{Step 1: Converting BGR (Blue-Green-Red) Image to RGB
(Red-Green-Blue)}\label{step-1-converting-bgr-blue-green-red-image-to-rgb-red-green-blue}}

BGR and RGB are conventions for the order of the different colour
channels. They are not colour spaces. When converting BGR image to RGB,
there are no computations other than switching around the order. Different
image processing libraries have different pixel ordering. To
be compatible with other libraries, we convert the BGR image into RGB
format. For example, when we read an image using the OpenCV library in
Python by default it interprets BGR format, but when we plot the image
it takes the RGB format in matplotlib package in Python.

\hypertarget{step-2-grayscaling}{%
\subsection{Step 2: Grayscaling}\label{step-2-grayscaling}}

Gray-scaling is the process of converting an image to shades of
gray from other colour spaces like RGB. This helps to increase the
contrast and intensity of images \citep{8675114}. Gray
scale images require only one single byte for each pixel whereas colour
(RGB) image requires 3 bytes for each pixel. Hence, gray-scaling reduces
the dimension of an image.

Another advantage of using
gray-scaled images is to reduce model complexity. Consider an example of
a training neural article on RGB images of \(20 \times 20 \times 3\)
pixel. The input layer will have 1200 input nodes. Whereas for gray
scaled images the same neural network will need only 400
(\(20 \times 20\)) input node.

\hypertarget{step-3-gaussian-filtering-gaussian-blurringgaussian-smoothing}{%
\subsection{Step 3: Gaussian Filtering (Gaussian Blurring/Gaussian
Smoothing)}\label{step-3-gaussian-filtering-gaussian-blurringgaussian-smoothing}}

Gaussian smoothing is an image smoothing technique. Image
smoothing techniques are used to remove the noise that can be occurred
due to the source (camera sensor). Image smoothing techniques help
in smoothing images and removing intensity edges.

\begin{figure}[!ht]

{\centering \includegraphics[width=0.6\linewidth]{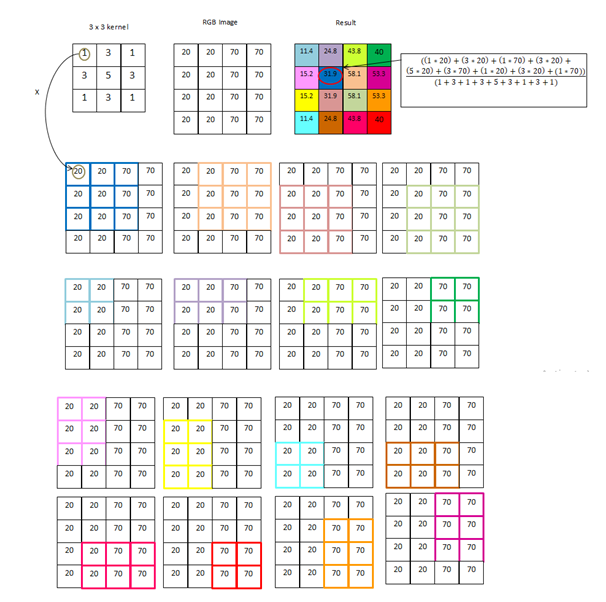} 

}

\caption{\label{fig:gau}Example of applying Gaussian smoothing}\label{fig:unnamed-chunk-3}
\end{figure}

A Gaussian function is used to blur the image. It is a linear filter. We used the OpenCV package in Python for Gaussian smoothing. The width and height of the Gaussian kernel must be positive and odd. Furthermore, the kernel standard deviation along the x and y-axis should be specified. When the kernel
standard deviation along the x-axis is specified, kernel standard deviation
along the y-axis is taken as equal to the kernel standard deviation
along the x-axis. However, if both kernel standard deviations are given as zeros,
the calculations are done using the kernel size. In our research, the width
and height of the kernel are defined as 55 and the kernel standard
deviation along the x-axis is assigned as zero. An example of applying
Gaussian smoothing to an image is shown in Figure \ref{fig:gau}.

\hypertarget{step-4-binary-thresholding}{%
\subsection{Step 4: Binary
Thresholding}\label{step-4-binary-thresholding}}

Thresholding is a segmentation technique that is used to separate
the foreground from its background. Thresholding converts the gray-scale images
into binary images according to the threshold value. If the pixel value is
smaller than the threshold value, the pixel value is set as 0, and if
not the pixel value is set to a maximum value which is generally 255.
The thresholding technique is done on the gray-scale images in computer vision.

We used Otsu's binarization
adaptive thresholding after Gaussian filtering to convert color images
to binary images. The reason for using Otsu's binarization method
is, it automatically determines the optimal threshold value.\\

\textbf{Otsu's Thresholding}\\

Nobuyuki Otsu introduced Otsu's method which is defined for a
histogram of grayscale values of a histogram (\(ghist_I\)) of an input
image \(Im\). To segment an image \(Im\) into two subsets of pixels
Otsu's method calculates an optimal threshold \(\tau\). The number of
pixel locations of the grayscale image is defined as \(|\Omega|\). The
algorithm maximizes the variance \(\sigma^2\) between the two subsets
(Within-class-variance) to find the threshold \(\tau\). The variance
\(\sigma^2\) is defined as

\[\sigma^2 = P_1(\mu_1-\mu)^2 + P_2(\mu_2-\mu)^2 = P_1P_2(\mu_1-\mu_2)^2, \]

where \(\mu\) is the mean of the histogram, \(\mu_1\) and \(\mu_2\) are
the mean values of first and second subset respectively. The
corresponding class probabilities, \(P_1\) and \(P_2\) are defined as
follows,

\[P_1 = \frac{\sum_{i=0}^{u}ghist_I(i)}{|\Omega|} \text{, and } P_2 = \frac{\sum_{i=u+1}^{255}ghist_I(i)}{|\Omega|},\]

where \(u\) is the candidate threshold and the maximum gray level
(\(G_{max}\)) is assumed as 255. To find the optimal threshold \(\tau\) for
segmenting image \(Im\), all candidate thresholds are evaluated this
way. The algorithm of Otsu's method is defined as follows:

\begin{table}[!ht]
\centering
\begin{tabular}{l}
\hline
Create a histogram for the grayscale image \\ 
Set the histogram variance $S_{max} = 0$                                             \\ \hline
\textbf{while} $u < G_{max}$ \textbf{do}                                             \\ \hline
Compute $\sigma^2 = P_1*P_2(\mu_1-\mu_2)^2$                                             \\ 
\textbf{if} $\sigma^2 > S_{max}$ \textbf{then}                                             \\ \hline
$S_{max} = \sigma^2$                                             \\
$\tau = u$                                             \\ \hline
\textbf{end if}                                             \\ 
Set $u = u+1$                                             \\ \hline
\textbf{end while}                                             \\ \hline
\end{tabular}
\caption{Otsu's method}
\label{tab:ot}
\end{table}

\begin{figure}[!ht]

\centering
       \includegraphics[width=60mm, height=50mm]{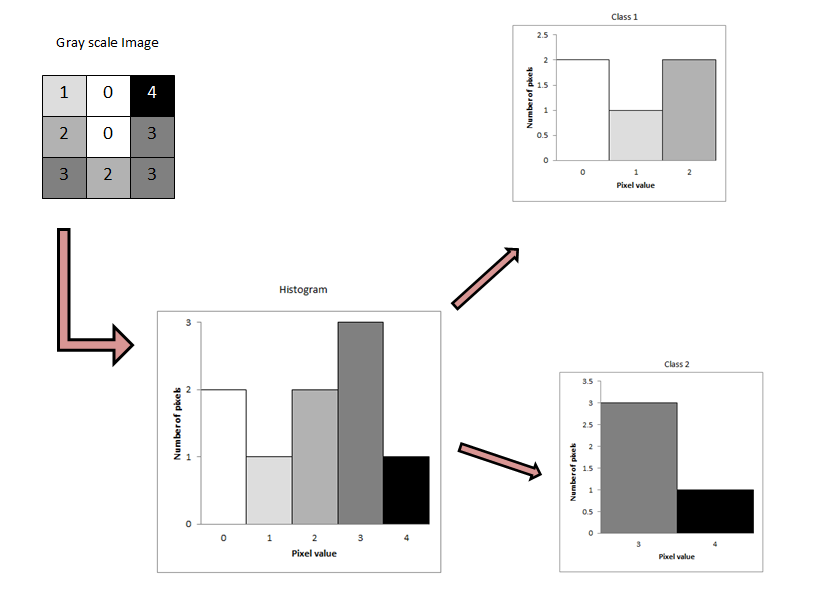}
        
        

\caption{Example of Otsu's binary thresholding}
\label{fig:otsu}
        \end{figure}

For example, assume that candidate threshold value \(u\) is 2.
Therefore, the image is separated into two classes, which are class 1
(\(\text{pixel value} <= 2\) ) and class 2 (\(\text{pixel value} > 2\)).
Class 1 represents the background and class 2 represents the foreground
of the grayscale image. According to Figure \ref{fig:otsu}, there are 9
pixel locations. The associated measurements are

\[P_1 = \frac{5}{9}, P_2 = \frac{4}{9}, \mu_1 = \frac{(0*2) +(1*1) + (2*2)}{(2+1+2)} = 1, \mu_2 = \frac{(3*3) +(4*1)}{(3+1)} = \frac{13}{4},\]

\[\sigma^2 = \frac{5}{9} * \frac{4}{9} * (1-\frac{13}{4})^2 = 1.25.\]

\hypertarget{step-5-image-resizing}{%
\subsection{Step 5: Image Resizing}\label{step-5-image-resizing}}

In this study we use two different benchmark leaf image datasets:
i) Flavia: 1907 fully color images of 32 classes of leaves \citep{wu2007leaf} and ii) Swedish: 1125 images from 15 different plant species
\citep{soderkvist2001computer}. Flavia and Swedish images have two different image sizes. To compare the results on different datasets, to improve the
memory storage capacity, and to reduce computational complexity the leaf
images are resized to a fixed resolution. In our study, the leaf images
have been resized to {[}1600 x 1200px{]} which is the size of Flavia
leaf images.

Other than the main image processing techniques discussed above,
the following techniques are applied to some images where necessary
after image thresholding.

`

\begin{figure}[!ht]
\begin{subfigure}{.5\textwidth}
\centering
        \includegraphics[width=50mm, height=30mm]{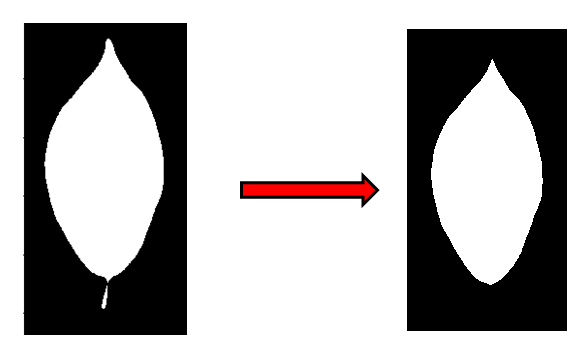}
        \caption{\label{fig:rst}Remove stalk}
        
\end{subfigure} 
\begin{subfigure}{.5\textwidth}
\centering
        \includegraphics[width=50mm, height=30mm]{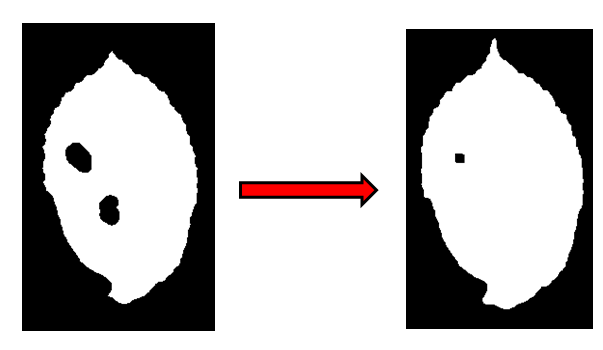}
        \caption{\label{fig:chl}Closing holes}
        
\end{subfigure} 

\caption{Other image processing steps: (a) Remove the stalk, and (b) close holes}
        \end{figure}

\hypertarget{remove-stalk}{%
\subsubsection{Remove the Stalk}\label{remove-stalk}}

As shown in Figure \ref{fig:rst}, this is used to remove the stalk
of the image. Remove the petiole (stalk) of the leaf image is another
version of the thresholding process. Thresholding is applied after finding
the sure foreground area. To find the sure foreground area, the distance
transform technique is used. The binary image is used as the input of
the distance transform technique. In the distance transform technique, an image is
created by assigning a number for each object pixel that corresponds to
the distance to the nearest background pixel. The distance is calculated
using the Euclidean distance. After finding
the sure foreground area, Otsu's binarization is applied again as the
thresholding technique. 

\hypertarget{closing-holes}{%
\subsubsection{Closing Holes}\label{closing-holes}}

Holes inside leaf areas occur due to plant disease, light
reflection, and noise. As shown in Figure \ref{fig:chl} closing holes is used to
remove small holes inside the foreground objects. This is achieved by
pulling background pixels to foreground pixels. The closing holes are also
known as dilation and while erosion does the opposite,  which is open
holes. This step is performed on a binary image.\\

\textbf{Erosion}\\

The basic idea of Erosion is that erodes the boundaries of
the foreground object. Since the input is a binary image, a pixel in the
original image is either 1 or 0. If all the pixels under the kernel is
1, a pixel of the original image is considered as 1, otherwise made to zero
(eroded). This means that depending upon the size of the kernel all
pixels near the boundary will be discarded. Therefore the thickness or size
of the foreground object decreases (the white region of the image
decreases).\\

\textbf{Dilation}\\

The opposite of erosion is defined as dilation. If at least one pixel
under the kernel is 1, the pixel element is 1 in Dilation. It tends to
increase the foreground of the image or the white region of the object.

\hypertarget{leaf-image-features}{%
\section{Leaf Image Features}\label{leaf-image-features}}

In the identification of plant species using leaf images, features of
leaves play an important role, because each leaf possesses a unique feature
that it makes different from others. The previous studies \citep{articlepl, article7, sun2017deep} use deep
learning neural networks to classify medicinal plants based on
pixel-based images. Given the leaf images, deep learning models
automatically identify features based on the pixel space of the images.
While deep learning models have achieved great success, the lack of
interpretability of features limits their widespread application. To
overcome this, we explore the use of interpretable, measurable, and
computer-aided features extracted from plant leaf images. We identified
52 features. The features are classified into four groups as i)
shape-based features, ii) color-based features, iii) texture-based
features and iv) scagnostic features. In this study, we considered 21 shape features, 4 texture features, 6 color-based features, and 21 scagnostic features.

\hypertarget{shape-features}{%
\subsection{Shape Features}\label{shape-features}}

\begin{figure}[!ht]

{\centering \includegraphics[width=0.7\linewidth]{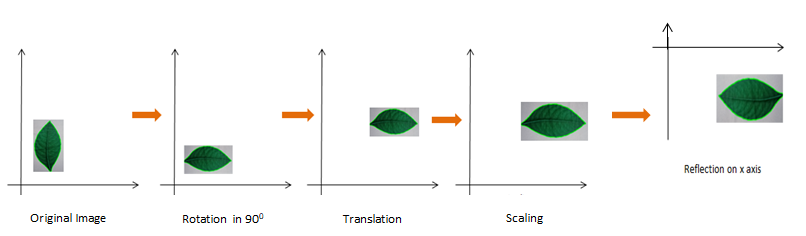} 

}

\caption{\label{img3} Illustration of geometric transformation}\label{fig:unnamed-chunk-5}
\end{figure}

When identifying real-world objects, the shape is known as an
essential sign for humans. We use the shape descriptors introduced by
\citep{articlee}. In addition to that, we introduce several
new shape features such as the number of convex points, x and y coordinates
of the center, number of maximum and minimum points, correlation of
Cartesian contour points, etc. The shape features should be invariant to
a certain class of geometric transformation of the object. The main
geometric transformations are rotation reflection, scaling, and
translation (see Figure \ref{img3}). The shape features we considered in
this study are invariant to the rotation and reflection. All the shape
features are extracted from the binary images.

\begin{figure}[!ht]

{\centering \includegraphics[width=0.3\linewidth]{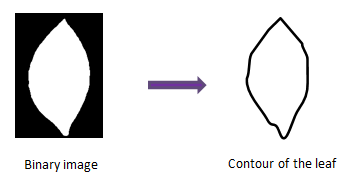} 

}

\caption{\label{cnt}Extract contour of the leaf image}\label{fig:unnamed-chunk-6}
\end{figure}

Extraction of image contour plays an important role in measuring the
shape of an image. Simply contour (see Figure \ref{cnt}) is a curve
joining all the continuous points (along the boundary), having the same
color or intensity.

\begin{figure}[!ht]

{\centering \subfloat[ \label{fig:trans-1}]{\includegraphics[width=0.1\linewidth]{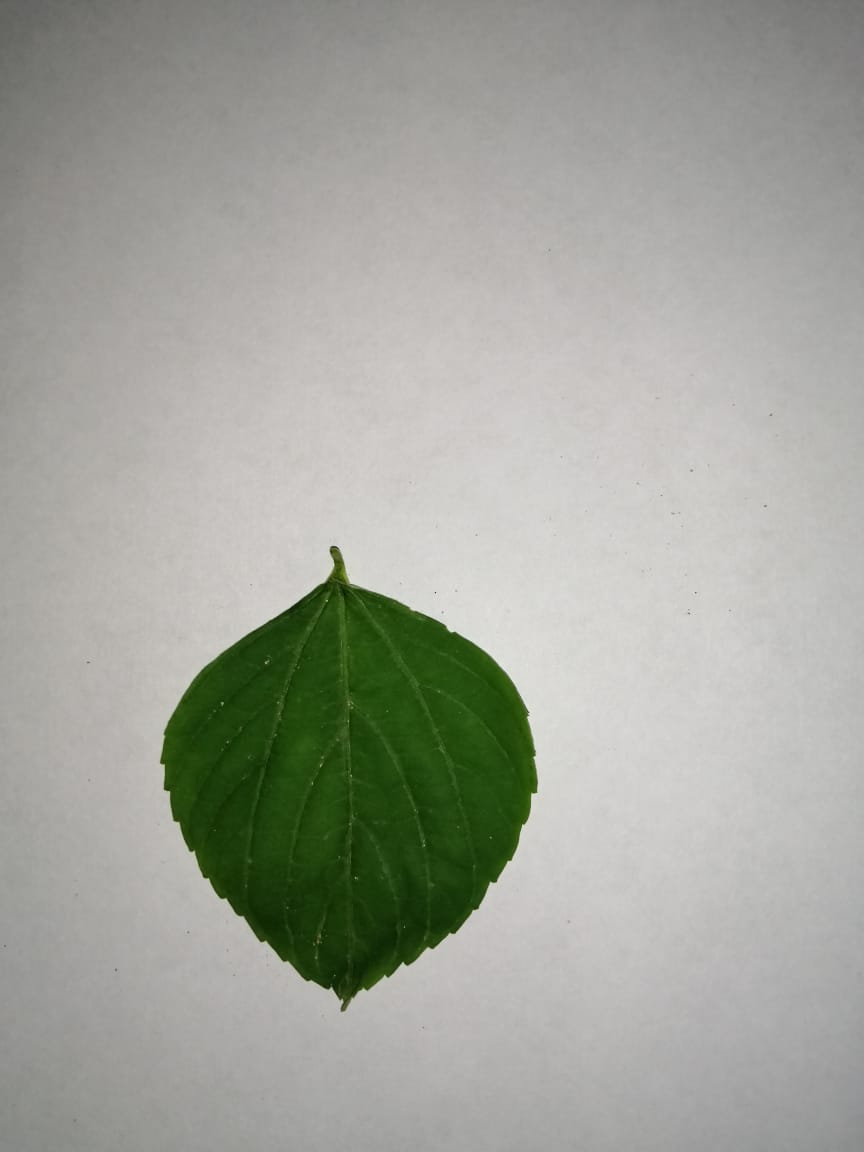} }\subfloat[\label{fig:trans-2}]{\includegraphics[width=0.1\linewidth]{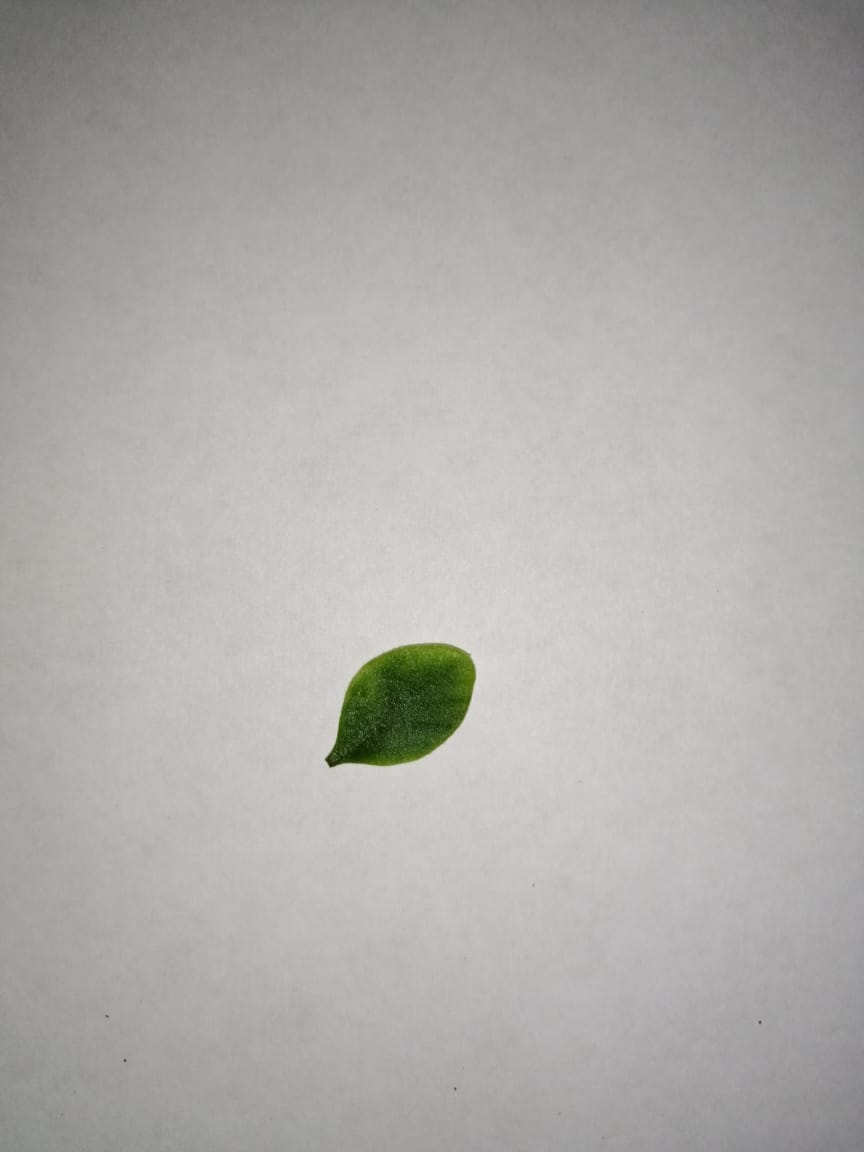} }

}

\caption{\label{trans} (a) Inappropriate translation,  (b) Inappropriate scaling }\label{fig:trans}
\end{figure}

To extract the contour of the leaf, the leaf should be placed
properly in the center of a white paper. If the image is placed as shown
in Figure \ref{fig:trans}, as a result of inappropriate translation (a)
and inappropriate scaling (b), problems arise in the calculations of
the contour. Furthermore, it is difficult to recognize the contour, when
the image is too small.

\begin{figure}[!ht]

{\centering \includegraphics[width=0.6\linewidth]{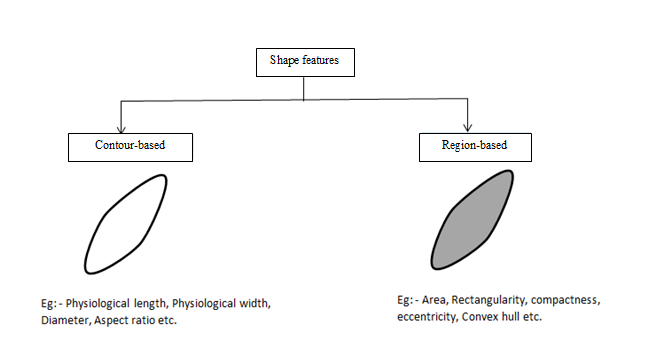} 

}

\caption{\label{scalimg4}Categorization of shape features}\label{fig:unnamed-chunk-7}
\end{figure}

Shape features can be classified into two main categories as
contour-based and region-based features \citep{articlee}. As
illustrated in \citep{articlee}, contour-based shape features
are computed based on the contour of a shape, whereas region-based shape
features are extracted from the whole region of a shape (see Figure
\ref{scalimg4}).

In this study, we use 6 initial shape features that are used to
derive 12 shape features. The 6 initial features are i) diameter, ii)
physiological length, iii) physiological width, iv) area, v) perimeter,
and vi) eccentricity.

\hypertarget{diameter-f_1}{%
\subsubsection{\texorpdfstring{Diameter
(\(F_1\))}{Diameter (F\_1)}}\label{diameter-f_1}}

Diameter is defined as the longest distance between any two points
on the margin of the leaf \citep{articlee}. To calculate the
diameter of the leaf image, first, we need to find the contour of the
leaf image. Then we need to select all pairs of contour points and
measure the Euclidean distance between the
two points separately. Finally, we have to find the maximum distance
among the calculated distances.







\hypertarget{physiological-length-f_2-and-physiological-width-f_3}{%
\subsubsection{\texorpdfstring{Physiological length (\(F_2\)) and
Physiological width
(\(F_3\))}{Physiological length (F\_2) and Physiological width (F\_3)}}\label{physiological-length-f_2-and-physiological-width-f_3}}

\begin{figure}[!ht]

{\centering \includegraphics[width=0.3\linewidth]{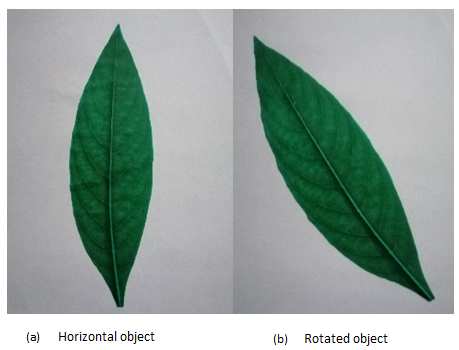} 

}

\caption{\label{act3}(a) Straight (horizontal or vertical) and (b) Rotated leaf image }\label{fig:act3}
\end{figure}

According to the authors in \citet{articlepl}, the physiological
length is ``measured based on the main vein of the leaf, as it stretches
from the main vein to the end tip.'' We use the definition by \citet{articlepl}, the physiological width is ``the span of leaf viewed from
one side to the other, from the leftmost point to the rightmost point of
leaf''. There are straight (horizontal or vertical) and angled leaf
images in our datasets, Flavia and Swedish (see example Figure
\ref{fig:act3}). There are two types of bounding rectangles.

\begin{enumerate}[noitemsep,nolistsep]
\def\labelenumi{\roman{enumi})}
\tightlist
\item
  Straight bounding rectangle: This is a straight rectangle that does not consider the rotation of the object (see Figure \ref{fig:act3}
  (a)).
\end{enumerate}

\begin{enumerate}[noitemsep,nolistsep]
\def\labelenumi{\roman{enumi})}
\setcounter{enumi}{1}
\tightlist
\item
  Rotated rectangle: This bounding rectangle is drawn with minimum
  background area. Therefore the rotation of the object is also
  considered (see Figure \ref{fig:act3} (b)).
\end{enumerate}

\begin{figure}[!ht]

{\centering \subfloat[\label{fig:bound-1}]{\includegraphics[width=0.3\linewidth]{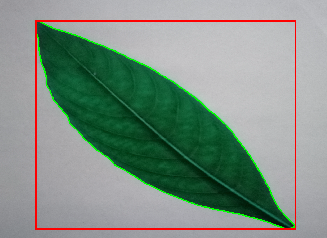} }\subfloat[\label{fig:bound-2}]{\includegraphics[width=0.3\linewidth]{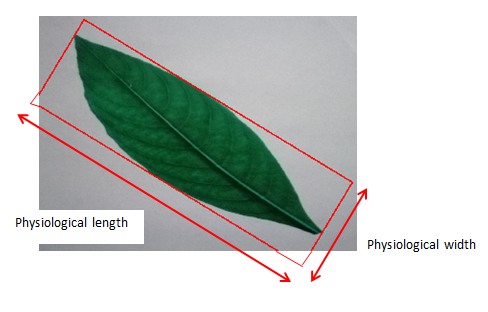} }

}

\caption{\label{bound} (a) Straight bounded rectangle of a rotated leaf image, (b) Rotated rectangle of angled leaf image (b)}\label{fig:bound}
\end{figure}

The straight bounding rectangle is enough to extract the physiological
length and physiological width of straight (horizontal or vertical) leaf
images. However, the straight bounding rectangle is not suitable to
compute the physiological length and physiological width of angled leaf
images. To solve this problem, we considered a rotated rectangle rather
than a bounded rectangle in computing shape features of angled images. As
shown in Figure \ref{bound}, the length of the rotated rectangle is considered as the physiological length ($F_2$) and the width of the rotated rectangle is considered as the physiological width ($F_3$).

\hypertarget{area-f_4}{%
\subsubsection{\texorpdfstring{Area
(\(F_4\))}{Area (F\_4)}}\label{area-f_4}}

\begin{figure}[!ht]
\begin{subfigure}{.5\textwidth}

\centering
        \includegraphics[width=50mm, height=40mm]{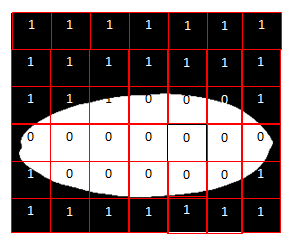}
        \caption{\label{areacal}Measure the area}
        
\centering
        \includegraphics[width=40mm, height=40mm]{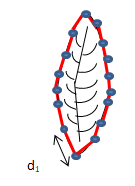}
        \caption{\label{calperi}Perimeter}

\end{subfigure} 
\begin{subfigure}{.5\textwidth}
\centering
        \includegraphics[width=60mm, height=50mm]{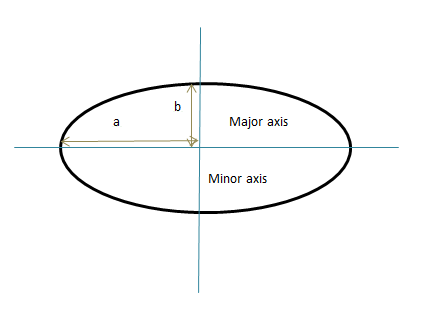}
        \caption{\label{shape6}Ellipse}
        
\centering
        \includegraphics[width=50mm, height=40mm]{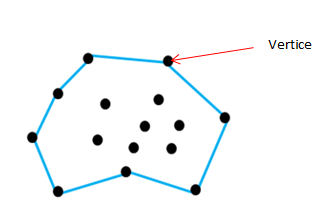}
        \caption{\label{shconvex}Convex hull}       
        
\end{subfigure} 

\caption{Illustration of (a) area, (b) perimeter, (c) ellipse, and (d) convex hull}
        \end{figure}

The area is computed after applying the thresholding process. Next, we
need to extract the best contour, and based on that contour area is
measured. The number of 0 pixels covered by the contour is the measure of
the area of the leaf image  \citep{articlepl}.

\hypertarget{perimeter-f_5}{%
\subsubsection{\texorpdfstring{Perimeter
(\(F_5\))}{Perimeter (F\_5)}}\label{perimeter-f_5}}

As shown in Figure \ref{calperi} perimeter is defined as the
summation of Euclidean distance of all continuous neighboring points in the contour.
Let \(n\) be the number of distances around the contour, then the
perimeter is defined as

\begin{equation}
   F_5 =  \sum_{i=0}^{n}d_i.
\label{equa_F5}
\end{equation}

\hypertarget{eccentricity-f_6}{%
\subsubsection{\texorpdfstring{Eccentricity
(\(F_6\))}{Eccentricity (F\_6)}}\label{eccentricity-f_6}}

Eccentricity is a characteristic of any conic section of a leaf
\citep{articlepl}. Eccentricity is defined that how much the ellipse
actually varying being circular. Eccentricity is calculated using
the equation \ref{eqecc} as

\begin{equation}
    F_6 = \sqrt{1-\frac{b^2}{a^2}},
    \label{eqecc}
\end{equation}

where \(a\) is semi-major axis and \(b\) is semi-minor axis (see Figure
\ref{shape6}).

\begin{figure}[!ht]
\begin{subfigure}{.5\textwidth}

\centering
        \includegraphics[width=40mm, height=25mm]{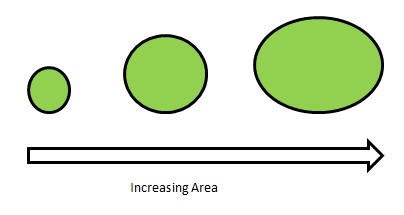}
        \caption{\label{areshape1}Area}

\centering
        \includegraphics[width=40mm, height=25mm]{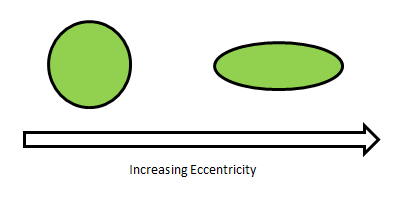}
        \caption{\label{shape5}Eccentricity}

\centering
        \includegraphics[width=40mm, height=25mm]{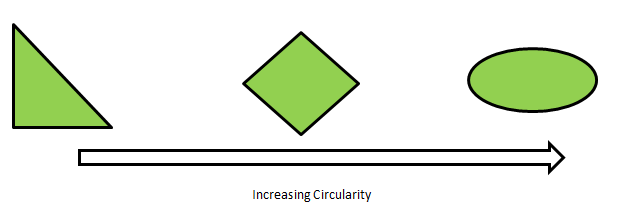}
        \caption{\label{shape3}Circularity}
        
\end{subfigure} 
\begin{subfigure}{.5\textwidth}

\centering
        \includegraphics[width=40mm, height=25mm]{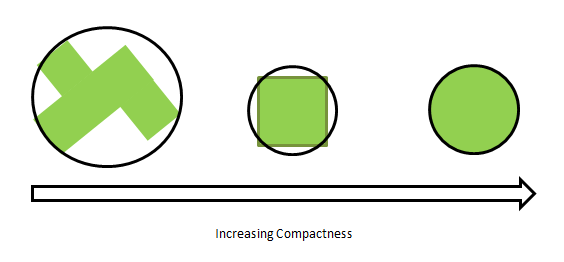}
        \caption{\label{shape4}Compactness}

\centering
        \includegraphics[width=40mm, height=30mm]{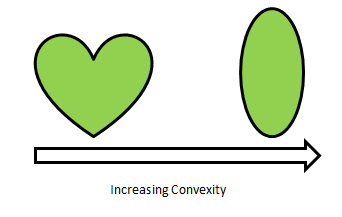}
        \caption{\label{shape7}Convexity}
        
\end{subfigure} 

\caption{Variations in shapes as the values of features increase.}
        \end{figure}

The eccentricity of an ellipse is varied between 0 and 1. If
eccentricity is 0, then we obtain a circle whereas eccentricity is 1,
then we obtain an ellipse (see Figure \ref{shape5}).

\hypertarget{number-of-convex-points-f_21-and-verticies}{%
\subsubsection{\texorpdfstring{Number of Convex Points (\(F_{21}\)) and
Verticies}{Number of Convex Points (F\_\{21\}) and Verticies}}\label{number-of-convex-points-f_21-and-verticies}}

\citet{inproceedings44} defined a convex hull as a
``hull that contains all the straight line segments connecting any pair
of points in its interior.'' The convex hull bounds a single polygon
(see Figure \ref{shconvex}). We introduced two new features computed
based on the convex hull: i) Number of convex points (\(F_{21}\)) and
ii) Number of convex vertices (\(F_7\)).

\hypertarget{roundness-circularity}{%
\subsubsection{Roundness/ Circularity}\label{roundness-circularity}}

Roundness is named as aka form factor, circularity, or
isoperimetrical factor. Roundness illustrates the difference between the
leaf and a circle. Equation \ref{calround} is used to calculate
roundness. Figure \ref{shape3} visual explanation for roundness measure.
The roundness is measured by

\begin{equation}
    R = \frac{4 \pi F_4}{{F_5}^2}.
\label{calround}
\end{equation}

\hypertarget{compactness}{%
\subsubsection{Compactness}\label{compactness}}

Compactness is closely related to roundness. Compactness measures
that how compatible the leaf fits a circle area (see Figure
\ref{shape4}). The compactness is measured using

\begin{equation}
    C = \frac{{F_5}^2}{F_4}.
\label{calcompact}
\end{equation}

\hypertarget{convexity}{%
\subsubsection{Convexity}\label{convexity}}

Convexity measures the curvature of the convex hull.

The equations and software packages used to compute shape features are
shown in Table \ref{tab:table1}.

\begin{longtable}{p{1cm}lllll}

\hline
Shape feature        & \multicolumn{1}{c}{Feature name}                                                            & \multicolumn{1}{c}{Figure} & \multicolumn{1}{c}{Formula} & \multicolumn{1}{c}{Range} & \begin{tabular}[c]{@{}l@{}}Software \\ package\end{tabular}   \\ \hline
\endfirsthead
\multicolumn{6}{c}%
{{\bfseries Table \thetable\ continued from previous page}} \\
\hline
Shape feature        & \multicolumn{1}{c}{Feature name}                                                            & \multicolumn{1}{c}{Figure} & \multicolumn{1}{c}{Formula} & \multicolumn{1}{c}{Range} & \begin{tabular}[c]{@{}l@{}}Software \\ package\end{tabular}   \\ \hline
\endhead
\hline
\endfoot
\endlastfoot
$F_1$  & Diameter                                                                                    &  \centering\includegraphics[width=10mm, height=15mm]{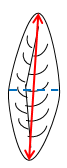}                          & \multicolumn{1}{l}{see definition $F_1$}        & \multicolumn{1}{l}{[0,$\infty$]}      & \begin{tabular}[c]{@{}l@{}}combinations,\\ numpy\end{tabular} \\
$F_2$                     & Physiological length                                                                        &     \centering\includegraphics[width=10mm, height=15mm]{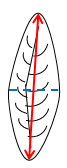}                       &            see definition $F_2$                &         [0,$\infty$)                 & OpenCV                                                        \\
$F_3$                     & Physiological width                                                                         &    \centering\includegraphics[width=10mm, height=15mm]{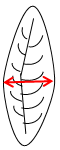}                        &     see definition $F_3$                       &           [0,$\infty$)                & OpenCV                                                        \\
$F_4$                     & Area                                                                                        &     \centering\includegraphics[width=10mm, height=15mm]{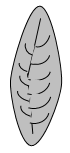}                        &       see definition $F_4$                   &                           & OpenCV                                                        \\
$F_5$                     & Perimeter                                                                                   &      \centering\includegraphics[width=10mm, height=15mm]{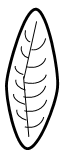}                      &        see definition $F_5$                    &          [0,$\infty$)                 & OpenCV                                                        \\
$F_6$                     & Eccentricity                                                                                &                            &      see definition $F_6$                       & [0,1]                     & OpenCV                                                        \\
$F_7$, $F_8$                     & \begin{tabular}[c]{@{}l@{}}x and y coordinate \\ of center\end{tabular}                     &      \centering\includegraphics[width=10mm, height=15mm]{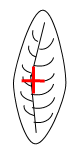}                      &                             &                           &            scipy.ndimage                                                   \\
$F_{9}$                     & Aspect ratio                                                                                &     \centering\includegraphics[width=10mm, height=15mm]{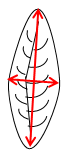}                       &       $F_9 = \frac{F_2}{F_3}$                      &       [0,$\infty$)                    &                                -                               \\
$F_{10}$                     & Roundness/ Circularity                                                                      &       \centering\includegraphics[width=10mm, height=15mm]{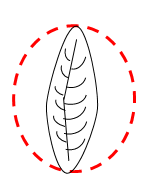}                       &           eq:\ref{calround}                  &          [0,$\infty$)                &     numpy                                                          \\
$F_{11}$                     & Compactness                                                                                 &     \centering\includegraphics[width=10mm, height=15mm]{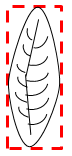}                       &        eg:\ref{calcompact}                     &     (0,$\infty$)                      &                   -                                            \\
$F_{12}$                     & Rectangularity                                                                              &                            &       $F_{12} = \frac{{F_5}^2}{F_4}$                      &      (0,$\infty$)                     &               -                                                \\
$F_{13}$                     & Narrow factor                                                                               &       \centering\includegraphics[width=10mm, height=15mm]{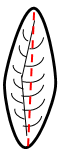}                     &         $F_{13} = \frac{F_1}{F_2}$                    &     [0,$\infty$)                      &                     -                                          \\
$F_{14}$                     & Perimeter ratio of diameter                                                                 &       \centering\includegraphics[width=10mm, height=15mm]{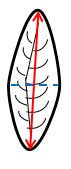}                     &         $F_{14} = \frac{F_5}{F_1}$                     &       [0,$\infty$)                    &                      -                                         \\
$F_{15}$  & \begin{tabular}[c]{@{}l@{}}Perimeter ratio \\ of physiological length\end{tabular}          &    \centering\includegraphics[width=10mm, height=15mm]{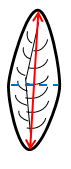}                        & $F_{15} = \frac{F_5}{F_2}$        & \multicolumn{1}{c}{[0,$\infty$)}      &         -                                                      \\
$F_{16}$  & \begin{tabular}[c]{@{}l@{}}Perimeter ratio of\\ physiological length and width\end{tabular} &    \centering\includegraphics[width=10mm, height=15mm]{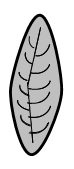}                        & $F_{16} = \frac{F_5}{F_2 * F_3}$        & \multicolumn{1}{c}{[0,$\infty$)}      &           -                                                    \\
$F_{17}$ & Perimeter convexity                                                                         &        \centering\includegraphics[width=10mm, height=15mm]{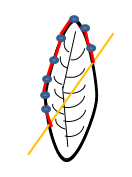}                    & \multicolumn{1}{c}{$F_{17} = \frac{\text{Perimeter of convex hull}}{F_5}$}        & \multicolumn{1}{c}{[0,$\infty$)}      &           OpenCV                                                    \\
$F_{18}$  & Area convexity                                                                              &    \centering\includegraphics[width=10mm, height=15mm]{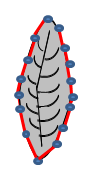}                         & \multicolumn{1}{c}{$F_{18} = \frac{(\text{Area of convex hull}-F_4)}{F_4}$}        & \multicolumn{1}{c}{[0,$\infty$)}      &               OpenCV                                                \\
$F_{19}$  & Area ratio of convexity                                                                     &     \centering\includegraphics[width=10mm, height=15mm]{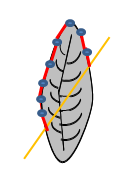}                        & $F_{19} = \frac{F_4}{\text{Area of convex hull}}$        & \multicolumn{1}{c}{[0,$\infty$)}      &               OpenCV                                                \\
$F_{20}$  & Equivalent diameter                                                                         &    \centering\includegraphics[width=10mm, height=15mm]{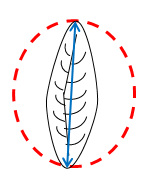}                        & $F_{20} = \sqrt{\frac{4*F_4}{\pi}}$       & \multicolumn{1}{c}{[0,$\infty$)}      &      numpy                                                         \\
\multicolumn{1}{l}{$F_{21}$} & \begin{tabular}[c]{@{}l@{}}Number of \\ convex points\end{tabular}                          &    \centering\includegraphics[width=10mm, height=15mm]{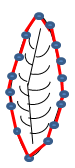}                        & number of convex points        & \multicolumn{1}{c}{[0,$\infty$)}      & OpenCV                                                        \\ \hline
\caption{Summary of shape features.}
\label{tab:table1}\\
\end{longtable}

\hypertarget{texture-features}{%
\subsection{Texture Features}\label{texture-features}}

Texture features are used to describe the surface or the
appearance of the leaf image. Texture can be assessed using a group of
pixels. Color is a property of a pixel. The texture is defined as a feeling of
various materials to human touch and texture is quantified based on
visual interpretation of this feeling. The leaf surface is a natural texture
that has random persistent patterns and does not show detectable
quasi-periodic structure \citep{articlee}. Therefore to
describe the natural texture patterns of the leaf fractal theory
\citep{articlee} is the best approach.\\
\hspace*{0.333em}\hspace*{0.333em}\hspace*{0.333em}\hspace*{0.333em}\hspace*{0.333em}\hspace*{0.333em}The
In this study we used Haralick texture features \citep{article31}. Haralick texture features  are functions of the normalized
GLCM (Gray Level Co-occurrence Matrix) is a common method to
represent image texture.

\[GLCM = \begin{bmatrix}
h(1,1) & h(1,2)  & \cdot &\cdot &\cdot & h(1,n) \\ 
h(2,1) & h(2,2)  & \cdot &\cdot &\cdot & h(2,n) \\ 
\cdot  & \cdot & \cdot  & & &\cdot\\ 
\cdot  & \cdot &  & \cdot& &\cdot\\ 
\cdot  & \cdot &  & & \cdot &\cdot\\ 
h(n,1) & h(n,2) & \cdot &\cdot &\cdot & h(n,n)
\end{bmatrix}\]

The GLCM is square with dimension \(n\), where \(n\) is the number
of gray levels in the image. Let \(h(a,b)\) is the probability that a
pixel with value \(a\) will be found adjacent to a pixel of value \(b\).
Then \(h(a,b)\) is defined as

\[h(a,b) = \frac{\text{Number of times a pixel with value a is adjacent to a pixel with value b}}{\text{Total number of such comparisons made}}.\]

\begin{figure}[!ht]

{\centering \includegraphics[width=0.2\linewidth]{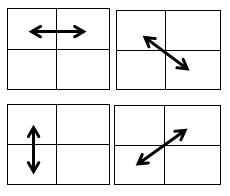} 

}

\caption{\label{direction}Four directions of adjacency as defined for calculation of the Haralick texture features}\label{fig:unnamed-chunk-9}
\end{figure}

\begin{figure}[!ht]

{\centering \includegraphics[width=0.5\linewidth]{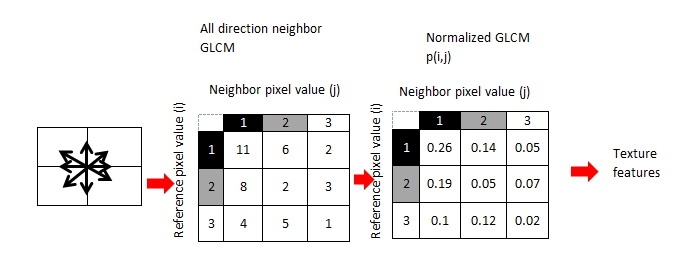} 

}

\caption{\label{img2}Computing the Haralick texture features from a 4 × 4 example image with all direction}\label{fig:unnamed-chunk-11}
\end{figure}

In order to calculate \(h(a,b)\), adjacency can be defined to
occur in each of four directions in a 2D (see Figure \ref{img2}), square
pixel image (horizontal, vertical, left and right diagonals - see
equation \ref{direction}).

\begin{figure}[!ht]

{\centering \includegraphics[width=0.5\linewidth]{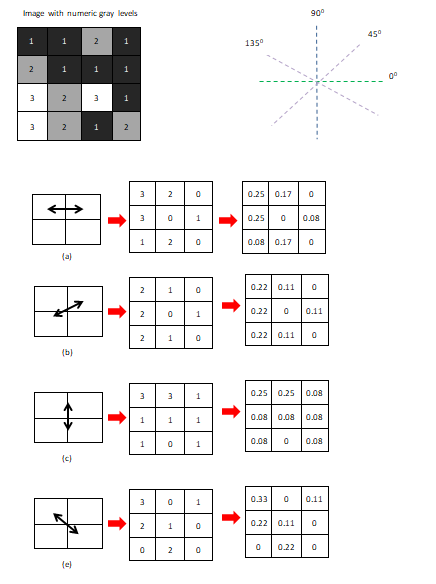} 

}

\caption{\label{img1}Computing the Haralick texture features from a 4 × 4 example image step by step}\label{fig:unnamed-chunk-10}
\end{figure}

The Haralick statistics are calculated based on the
matrices generated using each of these directions (see Figure
\ref{img1}) of adjacency . Haralick introduced 14 statistics to describe
the texture of the image based on the four co-occurrences matrices
generated \citep{article31}. In this research, we only used the following 4 statistics
among 14 of them, because most of the researchers used these 4
statistics as texture features (see Figure \ref{tabmytable}) of leaf
images. All the texture features are extracted from the grayscale
image. Texture features are calculated from the mahotas package in
Python. Table \ref{tabmytable} shows the definitions of texture
features.

\begin{table}[!ht]
\resizebox{\textwidth}{!}{%
\begin{tabular}{cclcc}
\hline
Texture feature & Feature name                                                             & \multicolumn{1}{c}{\begin{tabular}[c]{@{}c@{}}Detailed \\ description\end{tabular}}                                                                                                                       & Formula & Value range \\ \hline
    $F_{22}$            & Contrast                                                                 & \begin{tabular}[c]{@{}l@{}}Measures the\\ relation or difference\\ between the highest and\\ lowest gray levels of the GLCM\end{tabular}                                                                  &   $\frac{\sum_{a=1}^{columns}\sum_{b=1}^{rows}(a-b)^2 h(a,b)}{\text{Number of gray levels}-1}$      &     [0,$\infty$]        \\
     $F_{23}$           & Entropy                                                                  & \begin{tabular}[c]{@{}l@{}}Measures the randomness which means\\ that how uniform the image is\end{tabular}                                                                                               &    $-\sum_{a=1}^{columns}\sum_{b=1}^{rows}h(a,b)log_2(h(a,b))$     &    [$-\infty$,0]         \\
      $F_{24}$          & Correlation                                                              & \begin{tabular}[c]{@{}l@{}}Measurement of dependence\\ of gray levels of the GLCM. \\ It measures that how a particular\\ pixel is correlated to it's neighbor pixel\\ over the whole image\end{tabular} &  $\frac{\sum_{a=1}^{columns}\sum_{b=1}^{rows}(ab)h(a,b)-\mu_{x}\mu _{y}}{\sigma _{x}\sigma _{y}}$       &    [-1,1]         \\
      $F_{25}$          & \begin{tabular}[c]{@{}c@{}}Inverse \\ difference \\ moments\end{tabular} & \begin{tabular}[c]{@{}l@{}}It's a measure of homogeneity.\\ Inverse level of contrast that measures \\ how close the values of GLCM to\\ diagonal values in GLCM\end{tabular}                             &   $\sum_{a=1}^{columns}\sum_{b=1}^{rows}\frac{h(a,b)}{(a-b)^2}$      &     [0,$\infty$]        \\ \hline
\end{tabular}%
}
\caption{Definitions of texture features}
\label{tabmytable}
\end{table}

Let \(h(a, b)\) = Probability density function of gray - level pairs
\((a,b)\) and dimension of GLCM is \(n \times n\)
(\(\text{Number of columns} \times \text{Number of rows}\)). The
associated measures are given by

\[\mu_{x} = \sum_{a=1}^{columns}a\sum_{b=1}^{rows}h(a,b), \mu_{y} = \sum_{b=1}^{rows}b\sum_{a=1}^{columns}h(a,b),\]

\[\sigma_{x} = \sum_{a=1}^{columns}(a-\mu_{x})^2\sum_{b=1}^{rows}h(a,b), \sigma_{y} = \sum_{b=1}^{rows}(b-\mu_{y})^2\sum_{a=1}^{columns}h(a,b).\]

\hypertarget{color-features}{%
\subsection{Color Features}\label{color-features}}

\begin{figure}[!ht]

{\centering \includegraphics[width=0.3\linewidth]{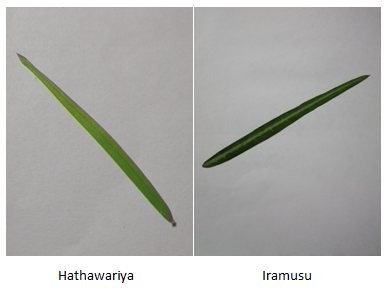} 

}

\caption{\label{leafimg} Images of plant species Hathawariya and Iramusu.}\label{fig:unnamed-chunk-12}
\end{figure}

Color is an important characteristic of images \citep{articlee,inproceedings1}. Some of the leaf images have very
similar shapes like Hathawariya (Figure \ref{leafimg}) and Iramusu
(Figure \ref{leafimg}). Even though shapes are similar in some leaves,
there are some differences in the colors of leaf images. Therefore in
addition to the shape features, we extract features related to the
color. The colors of an image are formed based on Red-Green-Blue (RGB)
colour channels of an image. Color properties are defined within a
particular color channel \citep{colarticle1,articlee}. In the field of image recognition, several general
color descriptors have been introduced. Color moments \citep{colarticle1,articlee} are the simple descriptor
among them. Mean, standard deviation skewness, and kurtosis are the
common moments. Color moments are convenient for real-time applications
due to their low dimension and low computational complexity.

We used mean (\(M\)) and standard deviation (\(SD\)) of intensity
values of red, green and blue channels as color features. Let \(h\) be
the number of pixels of the image and \(r\) is the channel type which
can be red, green, or blue. The corresponding colour features are
calculates as follows:

\begin{equation}
    M = \frac{\text{Total insensity value of } r^{th} \text{channel of the image pixels}}{\text{Total intensity value of the image}},
\label{equa2}
\end{equation}

\begin{equation}
    SD = \frac{\sqrt{\sum_{j=0}^{h}(r^{th} \text{channel intensity}_j - r^{th} \text{mean value})^2}}{\text{Total intensity value of the image}}.
\label{equa3}
\end{equation}

\hypertarget{scagnostic-features}{%
\subsection{Scagnostic features}\label{scagnostic-features}}

\begin{figure}[!ht]

\centering 
\includegraphics[width=0.7\linewidth]{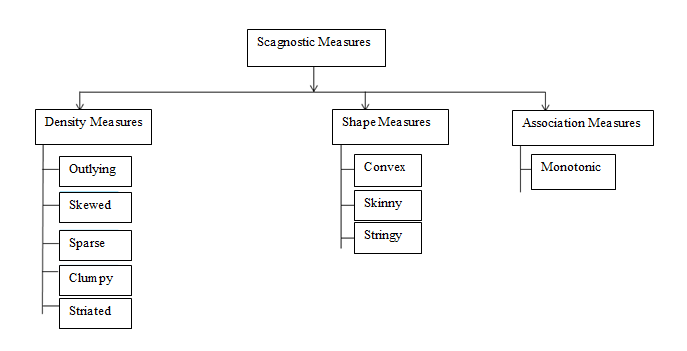} 

\caption{\label{scagimg}Hierarchy of Scagnostics}

\end{figure}

Scagnostic features are used to quantify the characteristics of 2D
scatterplot diagrams \citep{article37}. Scagnostic measures
are calculated based on the appearance of the scatterplot. All the
scagnostics features are calculated by using the R package called
\texttt{binostics}. Scagnostic features have the range of {[}0,1{]}.
There are 9 measures that are classified into three categories as shown
in Figure \ref{scagimg}. Based on binary images, scagnostic features are
extracted. To the best of our knowledge, this is the first time, the
scagnostic features are used for image recognition.

\begin{figure}[!ht]
        
\centering
        \includegraphics[width=60mm, height=50mm]{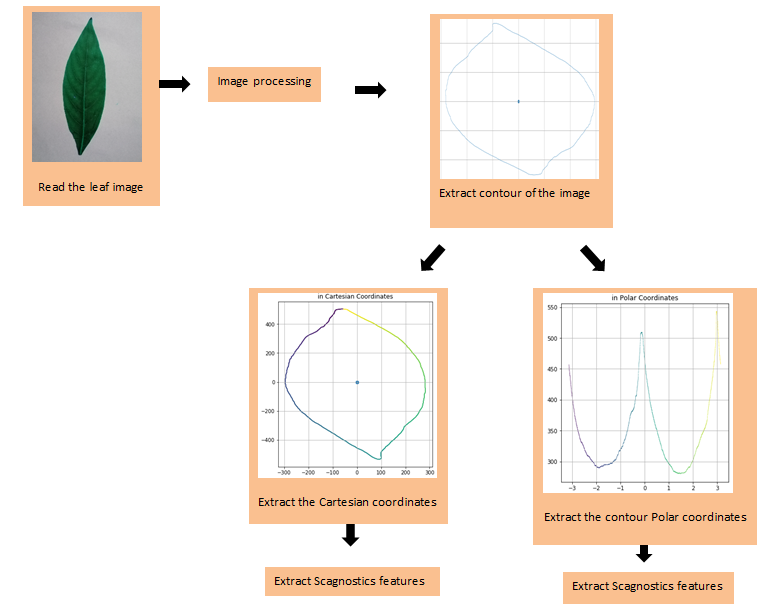}


\caption{\label{scp}Preprocessing for Scagnostics}
        \end{figure}

We separately measure the scagnostic features based on the cartesian
and polar coordinates of the contour (see Figure \ref{scp}). As the first
step, we have to extract the contour of the leaf image (see Figure
\ref{scp}). Then find the \(x\) and \(y\) coordinate values of the
cartesian and polar separately. The \(x\) and \(y\) coordinate values are
used to calculate the scagnostic features. The following definitions can
be useful in understanding scagnostics features.\\

\textbf{Geometric Graphs}

\begin{enumerate}
\def\labelenumi{\roman{enumi})}
\item
  Graph: A graph \(Gr = (Ve, Ed)\) is defined as a set of vertices
  (\(Ve\)) together with a relation on \(Ve\) induced by a set of edges
  (\(Ed\)). A pair of vertices is defined as an edge \(e(\nu,\omega)\),
  with e \(\in\) Ed and \(\nu,\omega \in\) Ve.
\item
  Geometric Graph: A geometric graph \(G* = [f(Ve), g(Ed), S]\); is an
  mapping of vertices to points and edges to straight lines to connect
  points in a metric space \(S\).
\end{enumerate}

\begin{enumerate}
\def\labelenumi{\roman{enumi})}
\setcounter{enumi}{2}
\item
  Length of an edge: The Euclidean distance between vertices that
  are connected to an edge is defined as the length of an edge, \(length(e)\).
\item
  Length of a graph: The sum of the lengths of edges in a graph is known
  as the length of a graph, \(length(Gr)\).
\item
  Path: A list of successively adjacent, distinct edges are known as a
  path. If the first and last vertices coincide, then the path is closed.
\item
  Polygon: A region bounded by a closed path is known as a polygon
  (\(P\)). A polygon bounded by exactly one closed path with no
  intersecting edges is known as a simple polygon.
\item
  The perimeter of a simple polygon: The length of the boundary of a simple
  polygon is known as the perimeter of a simple polygon. The area of
  the interior of a simple polygon is known as the area of a simple polygon.
\end{enumerate}

\begin{figure}[!ht]
\begin{subfigure}{.5\textwidth}
\centering
        \includegraphics[width=60mm, height=40mm]{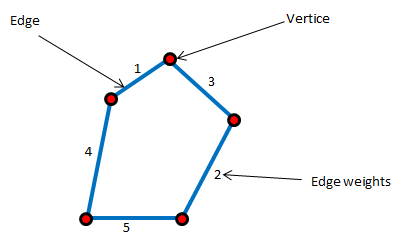}
        \caption{\label{scagimg5}Graph with 5 vertices and 5 edges}

\end{subfigure} 
\begin{subfigure}{.5\textwidth}
\centering
       \includegraphics[width=60mm, height=45mm]{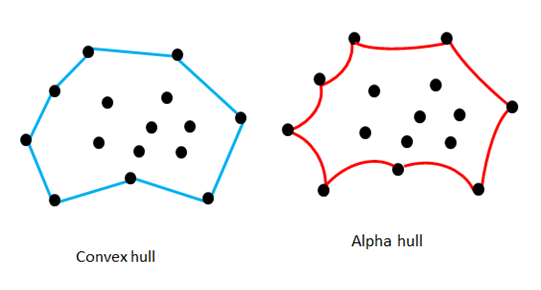}
        \caption{\label{scagimg4}Convex hull and alpha hull}
        
\end{subfigure} 

\caption{Illustration of vertices, edges, convex hull, and alpha hull}
        \end{figure}

The calculation of scagnostic features require identification of the
minimum spanning tree, convex hull, and alpha hull.\\

\textbf{Minimum Spanning Tree:}\\

The spanning tree of the graph is defined as \(G'(V', E')\), where
\(V' = Ve\), \(E' \subset Ed\) and \(E' = |Ve|-1\). A graph can have
more than one spanning tree. The spanning-tree should not be disconnected
and should not contain any cycles. A spanning tree whose total length is
least of all spanning trees on a given set of points is known as a
Minimum Spanning Tree (MST).\\

\textbf{Convex hull and Alpha hull}

\begin{enumerate}
\def\labelenumi{\roman{enumi})}
\tightlist
\item
  Convex Hull (see Figure \ref{scagimg4}): Given a set of points
  embedded in 2D Euclidean space, the convex hull of the set is the
  smallest convex polygon that contains all its points.
\end{enumerate}

\begin{enumerate}
\def\labelenumi{\roman{enumi})}
\setcounter{enumi}{1}
\tightlist
\item
  Alpha Hull (see Figure \ref{scagimg4}): The alpha hull is a
  generalization of the convex hull. It is defined as the union of all
  convex hulls of the input within balls of radius alpha. In other words,
  it is a set of piecewise linear simple curves in the Euclidean plane
  associated with a given set of points in the 2D Euclidean space. An open
  disk (\(D(r)\)) with radius \(r\) is used to define the indicator
  function to identify the associated points. If a point is on the
  boundary of \(D\) then \(D\) \textit{touches} a point and if a point
  is inside \(D\) then \(D\) \textit{contains} a point.
\end{enumerate}

\hypertarget{preprocessing-steps-related-to-scagnostic-features}{%
\subsubsection{Preprocessing steps related to scagnostic
features}\label{preprocessing-steps-related-to-scagnostic-features}}

To improve the performance of the algorithm and robustness of
the measures, preprocessing techniques such as binning and deleting
outliers are used before computing geometric graphs related features.

\begin{enumerate}
\def\labelenumi{\roman{enumi})}
\tightlist
\item
  Binning
\end{enumerate}

As the first step of binning, the data are normalized to the unit
interval. Then use a \(40 \times 40\) hexagonal grid to aggregate the
points in each scatterplot. We reduce the bin size by half and rebind
until no more than 250 non-empty cells. If there are more than 250
non-empty cells. When selecting the bins two important points to
consider are: i) efficiency (too many bins slow down calculations of the
geometric graphs) and ii) sensitivity (fewer bins obscure features in
the scatterplots). To improve the performance, hexagon binning is used.
To manage the problem of having too many points that start to overlap,
hexagon binning is used. The plots of hexagonal binning are density
rather than points. There are several reasons for using hexagon binning
instead of square binning on a 2D surface. Hexagons are more similar to
circles than squares. To keep scagnostics orientation-independent this
bias reduction is important.

The weight function is defined as

\begin{equation}
    \text{weight} = 0.7 + \frac{0.3}{1 + t^2},
    \label{w2}
\end{equation}

where \(t=\frac{n}{500}\) (\(n\) is the number of vertex).

If \(n > 2000\) then this function is fairly constant. By using
hex binning the shape and the parameters of the function are determined.
In computing sparse, skewed, and convex scagnostics this weight function
is used to adjust for bias.

\begin{enumerate}
\def\labelenumi{\roman{enumi})}
\setcounter{enumi}{1}
\tightlist
\item
  Deleting Outliers
\end{enumerate}

To improve the robustness of the scagnostics, deleting outliers can be
used. A vertex whose adjacent edges in the MST all have a weight
(length) greater than \(\omega\) is defined as an outlier in this
context. By considering nonparametric criteria for the simplicity and
Tukey's idea choose the following weight calculation is

\begin{equation}
    \text{weight} = qu_{75} + 1.5(qu_{75} - qu_{25}),
    \label{w1}
\end{equation}

where \(qu_{75}\) is the 75th percentile of the MST edge lengths and
\((qu_{75} - qu_{25})\) is the interquartile range of the edge lengths.

\begin{enumerate}
\def\labelenumi{\roman{enumi})}
\setcounter{enumi}{2}
\tightlist
\item
  Degree of a Vertex
\end{enumerate}

The degree of a vertex in an undirected graph is the number of
edges associated with the vertex. For example, vertices of degree 5
means there are 5 edges associated with each vertex (see Figure
\ref{scagimg5}).

\hypertarget{definitions-of-scagnostic-features}{%
\subsubsection{Definitions of Scagnostic
Features}\label{definitions-of-scagnostic-features}}

The definitions of scagnostic features are defined as follows. Our
notations are as follows: i) Convex hull (\(CH\)), ii) Alpha hull
(\(Al\)) and iii) Minimum spanning tree (\(MST\)). In the following
section, we give a brief description of the calculation of scagnostic
measures. For more details on rotation techniques, see \citet{article37}.

\hypertarget{density-measures}{%
\subsubsection{Density Measures}\label{density-measures}}

\begin{enumerate}
\def\labelenumi{\roman{enumi})}
\tightlist
\item
  Outlying: The outlying measure is calculated before deleting the
  outliers for the other measures. The outlying measure is
\end{enumerate}

\begin{equation}
      F_{sc1} = \frac{\text{Total length of edges adjacent to outlying points}}{\text{Total edge length of the MST}}.
\end{equation}

\begin{enumerate}
\def\labelenumi{\roman{enumi})}
\setcounter{enumi}{1}
\tightlist
\item
  Skewed: The skewed measure is the first measure of relative density, a relatively robust measure of skewness in the distribution
  of edge lengths. After adaptive binning skewed tends to decrease with
  \(n\). The skewed measure is calculated using the equations
\end{enumerate}

\begin{equation}
    F_{sc2} = 1-\text{weight}*(1-qu_{skew}), 
\end{equation}

~~~~where,

\begin{equation}
    qu_{skew} = \frac{qu_{90}-qu_{50}}{qu_{90}-qu_{10}}, 
\end{equation}

~~~~and the calculation of \(weight\) is given in equation \ref{w2}.

\begin{enumerate}
\def\labelenumi{\roman{enumi})}
\setcounter{enumi}{2}
\tightlist
\item
  Sparse: The second relative density measure is a sparse measure that
  measures whether points in a 2D scatterplot are confined to a lattice
  or a small number of locations on the plane. The sparse is measured by
\end{enumerate}

\begin{equation}
   F_{sc3} = \text{weight} *qu_{90},
\end{equation}

~~~~where the weight function is equation \ref{w2} and \(qu_{90}\) is
the 90th percentile of the distribution of edge lengths in the \(MST\).

\begin{enumerate}
\def\labelenumi{\roman{enumi})}
\setcounter{enumi}{3}
\tightlist
\item
  Clumpy: Clumpiness measures the number of small-scale structures in
  the 2D-scatter plot (Wilkinson and Wills 2008). In order to calculate
  this feature another measurement called \(T\) is needed which is
  calculated based on the RUNT graph \citep{hartigan1992runt}. The clumpy
  is measured by
\end{enumerate}

\begin{equation}
    F_{sc4} = max_j[1-\frac{max_k[length(e_k)]}{length(e_j)}].
\end{equation}

~~~~In the formula below the \(j\) value goes over the edges in \(MST\)
and \(k\) runs over all edges in the RUNT graph.

\begin{enumerate}
\def\labelenumi{\alph{enumi})}
\setcounter{enumi}{21}
\tightlist
\item
  Striated: Striated define the coherence in a set of points as the
  presence of relatively smooth paths in the minimum spanning tree. This
  measure is based on the number of adjacent edges whose cosine is less
  than minus 0.75. The stratified is measured by
\end{enumerate}

\begin{equation}
    F_{sc5} = \frac{1}{|Ve|}\sum_{\nu \in Ve^{(2)}}^{}I(\cos\theta_{e(\nu,a)e(\nu,b)}<-0.75),
\end{equation}

~~~~where \(Ve^{(2)} \subseteq Ve\) and \(I()\) be an indicator
function.

\hypertarget{scagnostic-based-shape-measures}{%
\subsubsection{Scagnostic-based Shape
Measures}\label{scagnostic-based-shape-measures}}

~~~~~~~Both topological and geometric aspects of the shape of a set of
scattered points is considered. As an example, a set of scattered points
on the plane appeared to be connected, convex, and so forth, want to know
under the shape measures. By definition scattered points are not like
this. Therefore to make inferences additional machinery (based on
geometric graphs) is needed. By measuring the aspects of the convex
hull, the alpha hull, and the minimum spanning tree is determined.

\begin{enumerate}
\def\labelenumi{\roman{enumi})}
\tightlist
\item
  Convex: The ratio of the area of the alpha hall (\(Al\)) and the area
  of the convex hull (\(CH\)) is the base of measuring convexity. The
  convex is measured by
\end{enumerate}

\begin{equation}
    F_{sc6} = \text{weight} \times \frac{\text{Area of alpha hull}}{\text{Area of convex hull}},
\end{equation}

~~~~~~~where the weight function is equation \ref{w2}.

\begin{enumerate}
\def\labelenumi{\roman{enumi})}
\setcounter{enumi}{1}
\tightlist
\item
  Skinny: Skinny is measured by using the corrected and normalized ratio
  of perimeter to the area of polygon measures. The skinny is measured by
\end{enumerate}

\begin{equation}
    F_{sc7} = 1- \frac{\sqrt{4 \times \pi \times \text{Area of alpha hull}}}{\text{Perimeter of alpha hull}}.
\end{equation}

~~~~~~~Furthermore

\[F_{sc7} = \Bigg\{^{0; \text{ if circle}}_{\text{Near } 1 ; \text{ if skinny}.}\]

\begin{enumerate}
\def\labelenumi{\roman{enumi})}
\setcounter{enumi}{2}
\tightlist
\item
  Stringy: A skinny shape with no branches is known as a stringy shape.
  By counting the vertices of degree 2 in the minimum spanning tree and
  comparing them to the overall number of vertices minus the number of
  single-degree vertices, a skinny measure is calculated. To adjust for
  negative skew in its conditional distribution of \(n\), cube the
  stringy measure. The stringy is measured by
\end{enumerate}

\begin{equation}
    F_{sc8} = \frac{|Ve^{(2)}|}{|Ve| - |Ve^{(1)}|},
\end{equation}

~~~~~~~where \(Ve\) is the number of vertices.

\hypertarget{association-measure}{%
\subsubsection{Association Measure}\label{association-measure}}

~~~~~~~Symmetric and relatively robust measures of the association are
interested. To assess the monotonicity in a scatter plot, the squared
  Spearman correlation coefficient is used. In calculating monotonicity,
  the squared value of the coefficient is considered to remove the
  distinction between positive and negative coefficients. The reason is,
  the researchers are more interested in how strong the relationship is
  rather than their direction (negative or positive).

\hypertarget{number-of-minimum-f_8-and-maximum-points-f_9}{%
\subsubsection{\texorpdfstring{Number of Minimum (\(F_8\)) and Maximum
Points
(\(F_9\))}{Number of Minimum (F\_8) and Maximum Points (F\_9)}}\label{number-of-minimum-f_8-and-maximum-points-f_9}}

~~~~~~Number of minimum, and maximum points are new measures that are
obtained from the polar coordinate of leaf contour. The number of global
maximum points  ($F_8$) and the number of global
minimum points  ($F_9$) are also considered as features.

\hypertarget{correlation-of-cartesian-contour-f_10}{%
\subsubsection{\texorpdfstring{Correlation of Cartesian Contour
(\(F_{10}\))}{Correlation of Cartesian Contour (F\_\{10\})}}\label{correlation-of-cartesian-contour-f_10}}

~~~~~~Correlation is another new feature computed based on the cartesian
contour. The measure is calculated as

\begin{equation}
   F_{10} =  \frac{\sum_{i=0}^{m}(x_i - \overline{\rm x})(y_i - \overline{\rm y})}{\sqrt{\sum_{i=0}^{m} (x_i - \overline{\rm x})^2 (y_i - \overline{\rm y})^2}},
\label{equa_F10}
\end{equation}

~~~where \((x_i,y_i)\) is the coordinate of cartesian contour and
\(m\) is the number of points in the cartesian contour

\hypertarget{empirical-application}{%
\section{Empirical Application}\label{empirical-application}}

\hypertarget{data-sets}{%
\subsection{Data sets}\label{data-sets}}

~~~~~~We use two publicly available datasets to demonstrate the
applications of features. They are i) Flavia leaf image dataset and ii)
Swedish leaf image dataset

\hypertarget{flavia-leaf-image-dataset}{%
\subsubsection{Flavia Leaf Image
Dataset}\label{flavia-leaf-image-dataset}}

~~~~~~The Flavia dataset contains 1907 leaf images. There are 32
different species and each has 50-77 images. Scanners and digital
cameras are used to acquire the leaf images on a plain background. The
isolated leaf images contain blades only, without a petiole. These leaf
images are collected from the most common plants in Yangtze, Delta,
China \citep{articlee}. Those leaves were sampled on the campus
of the Nanjing University and the Sun Yat-Sen arboretum, Nanking, China
\citep{articlee} available at
\url{https://sourceforge.net/projects/flavia/files/Leaf%2520Image%2520Dataset/}.

\begin{figure}[!ht]
\begin{subfigure}{.5\textwidth}
\centering
        \includegraphics[width=60mm, height=40mm]{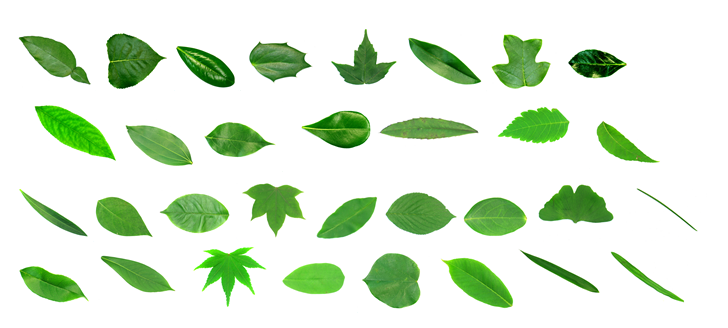}
        \caption{\label{slp1}Sample of Flavia dataset}
        
\end{subfigure} 
\begin{subfigure}{.5\textwidth}
\centering
        \includegraphics[width=60mm, height=40mm]{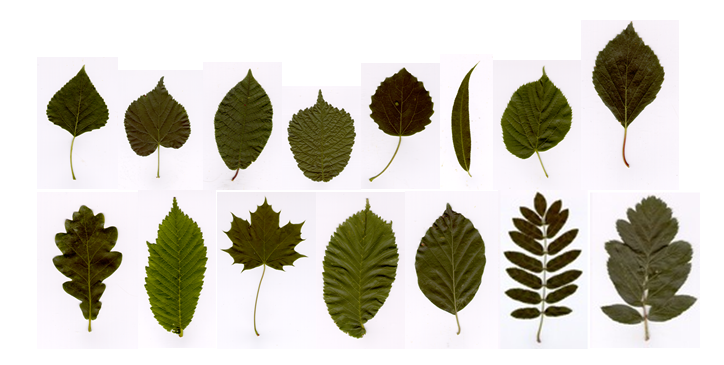}
        \caption{\label{slp2}Sample of Swedish leaf dataset}
        
\end{subfigure} 

\caption{Samples from Flavia and Swedish datasets}
        \end{figure}

\hypertarget{swedish-leaf-image-dataset}{%
\subsubsection{Swedish Leaf Image
Dataset}\label{swedish-leaf-image-dataset}}

The Swedish dataset contains 1125 images. The images of isolated
leaf scans on a plain background of 15 Swedish tree species, with 75
leaves per species. This dataset has been captured as part of a joined
leaf classification project between the Linkoping University and the
Swedish Museum of Natural History \citep{articlee} available at
\url{https://www.cvl.isy.liu.se/en/research/datasets/swedish-leaf/}.

We applied the aforementioned image processing techniques and computed
features from each image. 

\hypertarget{visualization-of-ability-of-features-to-distinguish-classes-of-interest}{%
\subsection{Visualization of ability of features to distinguish classes
of
interest}\label{visualization-of-ability-of-features-to-distinguish-classes-of-interest}}

In order to identify the ability of features to distinguish classes of
interest, we first label features according to their shapes as i)
diamond, ii) heart shape, iii) needle shape, iv) simple round, and v)
round shape (see Figure \ref{shapeimg}). These morphological
characteristics are identified by observing images in the medicinal
plant repository maintained by Barberyn Ayurveda resort and University
of Ruhuna available at \url{http://www.instituteofayurveda.org/plants/}.
Our observed results are converted into an open-source R software package
called MedLEA: \textbf{Med}icinal \textbf{LEAf} which is available on
Comprehensive R Archive Network \citep{medlea}.

We explore the ability of features to classify images under
supervised learning settings and unsupervised learning settings. For this
purpose, we use Linear Discriminant Analysis (LDA) and Principal
Component Analysis (PCA). LDA is a supervised dimensionality reduction
technique, and PCA is an unsupervised dimensionality reduction technique.
In this section, we visualize and compare the results obtained using
LDA, and PCA on Flavia and Swedish datasets. To compute the LDA projection
shape label is taken as the response variable. There are 5 main shape
categories as: diamond, simple round, round, needle, and heart shape.

We further explore the ability of our features to classify classes when there is a moderately large number of class labels. For that, we label the leaves according to their species type. In the Swedish data set, there are 12 species-wise class categories while the Flavia data set contains 32 species-wise class categories. Both PCA and LDA approaches were used to explore the class separability visually. To explore the class separability of species in the PCA space, we colour the points according to their species labels. For the corresponding LDA analysis, we use species class labels as the response variable.

\begin{figure}
\centering
\includegraphics[width=1\columnwidth]{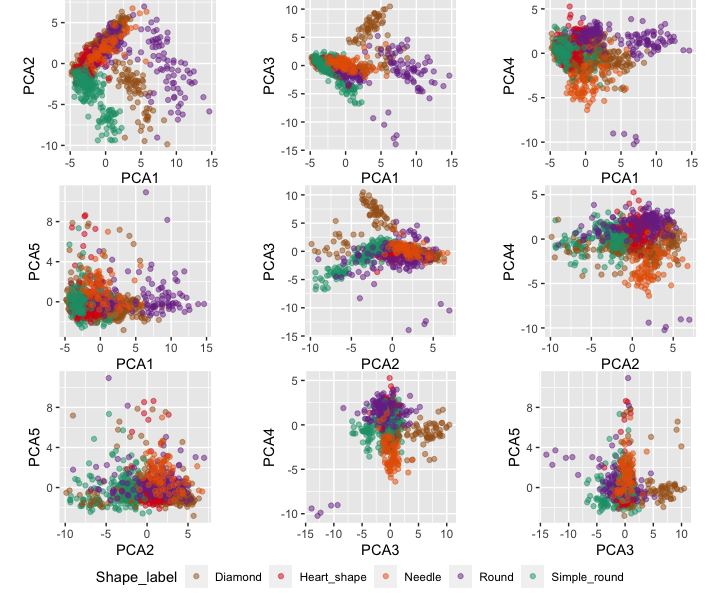}
\caption{\label{pcaswedish}Distribution of Swedish leaf images on the
principal component analysis-based projection space. All
projected points are coloured according to their shape labels.}
\includegraphics[width=1\columnwidth]{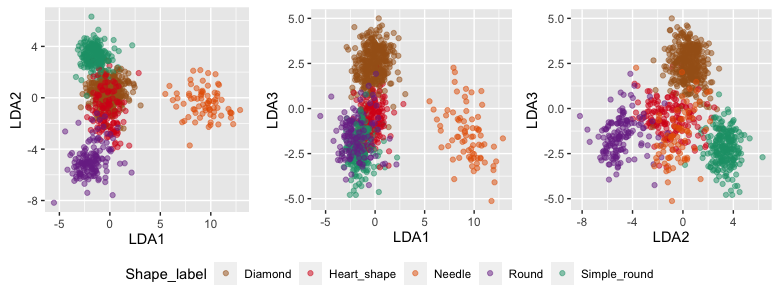}
\caption{\label{ldaswedish}Distribution of Swedish leaf images on linear discriminant analysis based projection space. All
projected points are coloured according to their shape labels.}
\end{figure}

\begin{figure}
\centering
\includegraphics[width=1\columnwidth]{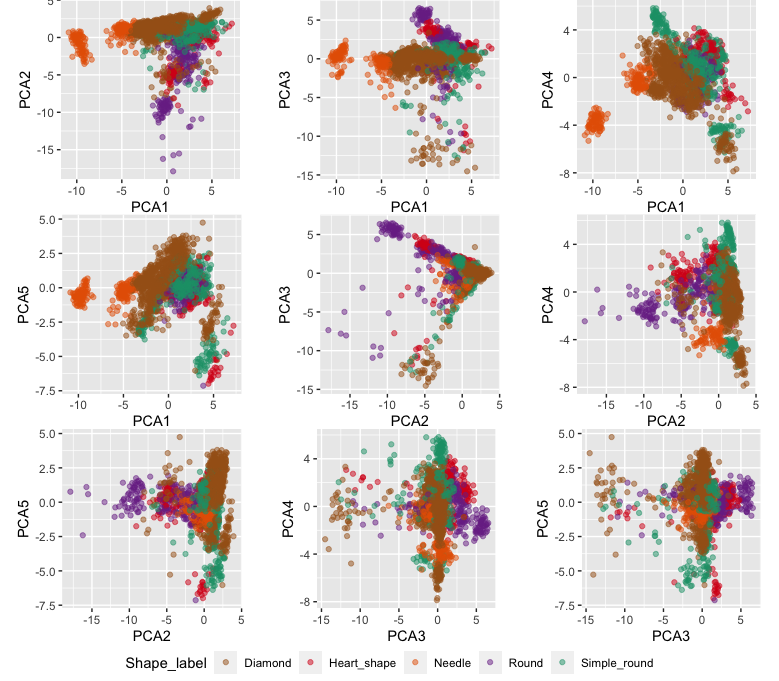}
\caption{\label{pcaflavia}Distribution of Flavia leaf images on the
principal component analysis based projection space. All
projected points are coloured according to their shape labels.}
\includegraphics[width=1\columnwidth]{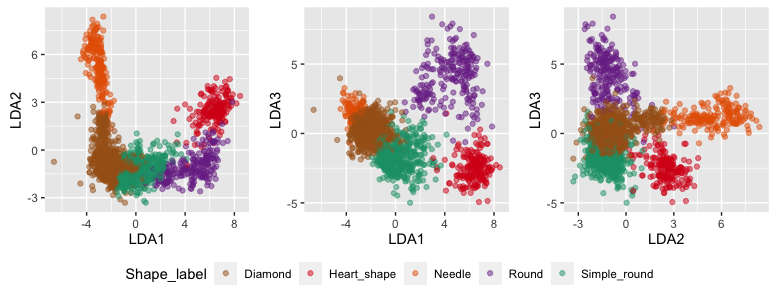}
\caption{\label{ldaflavia}Distribution of Flavia leaf images on linear discriminant analysis based projection space. All
projected points are coloured according to their shape labels.}
\end{figure}

\begin{table}[!ht]
\resizebox{\textwidth}{!}{%
\begin{tabular}{lrrrrrrrrrrrrrrrr}
\hline
\multirow{2}{*}{\begin{tabular}[c]{@{}l@{}}Feature\\ name\end{tabular}}     & \multicolumn{8}{c}{Swedish}                                                                                                   & \multicolumn{8}{c}{Flavia}                                                                                                      \\
                                                                            & PCA1         & PCA2         & PCA3         & PCA4          & PCA5         & LDA1           & LDA2           & LDA3            & PCA1         & PCA2          & PCA3          & PCA4         & PCA5          & LDA1           & LDA2           & LDA3            \\ \hline
Skewed polar                                                                & 0.05  & -0.08 & -0.11 & -0.03 & 0.03  & 1.48       & -0.49     & -0.17      & 0.02  & -0.03 & -0.10 & 0.09  & -0.02 & 0.01     & 1.96      & 0.60       \\
Clumpy polar                                                                & 0.04  & -0.07 & -0.07 & -0.01 & -0.03 & 5.65       & 2.42       & -5.47       & -0.02 & -0.03 & -0.06 & -0.01 & -0.02 & -1.46      & 1.25       & 1.85        \\
Sparse polar                                                                & 0.10  & -0.21 & -0.16 & -0.07 & -0.05 & -19.55      & -9.32      & 52.97        & 0.02  & -0.15 & -0.29 & 0.02  & 0.00 & -35.36      & 3.04       & 44.68        \\
Striated polar                                                              & -0.13 & 0.22  & 0.21  & 0.01  & -0.02 & -1.08      & -0.97     & 5.32        & -0.05 & 0.14  & 0.31  & -0.07 & 0.01  & 1.79       & 0.24      & 0.43       \\
Convex polar                                                                & 0.12  & -0.21 & -0.18 & -0.06 & -0.03 & 0.44      & 6.59       & 3.54        & 0.04  & -0.16 & -0.32 & 0.03  & 0.03  & 11.76       & -20.21      & 4.70        \\
Skinny polar                                                                & 0.01  & -0.09 & -0.04 & -0.02 & 0.08  & 0.67     & 0.06     & 0.19       & -0.01 & -0.02 & -0.08 & 0.05  & 0.00 & 0.13      & 0.04     & -0.12      \\
Stringy polar                                                               & -0.15 & 0.14  & 0.19  & -0.01 & -0.10 & -4.10      & 4.80       & -1.99       & -0.05 & 0.15  & 0.28  & -0.07 & 0.01  & -3.44      & -3.07      & 8.62        \\
Monotonic polar                                                             & 0.09  & 0.02  & -0.05 & 0.14  & 0.17  & -1.51      & -2.54      & 1.34        & 0.07  & 0.06  & 0.05  & -0.09 & 0.18  & 0.14      & 0.34      & -1.71       \\
Skewed contour                                                              & 0.07  & -0.16 & -0.15 & 0.01  & 0.03  & -1.99      & -2.33     & -2.28       & 0.01  & -0.04 & -0.09 & -0.05 & -0.06 & -0.39     & 0.10      & -2.35       \\
Clumpy contour                                                              & 0.04  & -0.05 & -0.08 & 0.05  & 0.05  & -8.59      & 3.43       & -0.49      & 0.02  & -0.02 & -0.04 & -0.01 & 0.00  & -1.07      & -2.38      & 1.29        \\
Sparse contour                                                              & 0.16  & -0.18 & -0.17 & -0.02 & 0.01  & 54.36       & 45.24       & 34.38        & 0.02  & -0.16 & -0.29 & 0.00 & 0.02  & -29.91      & 7.74       & -114.90       \\
Striated contour                                                            & -0.15 & 0.21  & 0.20  & 0.00 & -0.02 & 4.19       & 0.70      & 2.59        & -0.04 & 0.15  & 0.31  & 0.01  & 0.02  & -0.11     & 0.24      & -4.12       \\
Convex contour                                                              & 0.12 & -0.21 & -0.17 & -0.05 & -0.03 & 4.92       & -8.62     & 8.77        & 0.04  & 0.015  & -0.05 & 0.09  & 0.10  & -0.43     & 2.56       & 0.89       \\
Skinny contour                                                              & 0.04  & -0.10 & -0.07 & 0.04  & 0.03  & 0.63      & 0.24      & -0.34      & -0.03 & 0.00  & -0.13 & -0.11 & 0.04  & 0.13      & 0.67      & -0.20      \\
Stringy contour                                                             & -0.13 & 0.08  & 0.13  & -0.02 & -0.09 & -1.68      & -4.53      & -4.57       & -0.04 & 0.15  & 0.30  & -0.01 & -0.03 & -1.63      & 0.34      & 0.29       \\
Monotonic contour                                                           & 0.01  & 0.04  & 0.01  & -0.17 & 0.07  & 6.63       & -1.24      & -3.58       & 0.05  & 0.03  & 0.08  & 0.25  & 0.05  & 0.25      & 0.57      & -0.21      \\
No of max points                                                            & 0.08  & -0.22 & -0.20 & 0.00 & -0.04 & 0.00   & -0.08    & -0.02     & -0.02 & -0.15 & -0.23 & 0.05 & -0.03 & 0.06     & 0.04     & -0.06     \\
No of min points                                                            & 0.07  & -0.21 & -0.18 & -0.01 & -0.06 & -0.05    & 0.07     & -0.01     & -0.02 & -0.16 & -0.22 & -0.12 & -0.04 & 0.01    & 0.00    & 0.14       \\
diameter                                                                    & 0.01  & -0.08 & 0.11  & 0.45  & -0.12 & 0.05     & -0.03    & -0.04     & -0.03 & 0.15  & -0.08 & -0.28 & 0.29  & 0.02     & 0.02     & 0.02      \\
area                                                                        & -0.19 & -0.21 & 0.05  & 0.16  & -0.07 & -0.00 & 0.00 & 0.00  & 0.28  & 0.04  & 0.00  & -0.01 & 0.15 & 0.00 & 0.00 & -0.00 \\
perimeter                                                                   & 0.09  & -0.20 & 0.29  & 0.11  & -0.03 & -0.01    & -0.00   & -0.00 & 0.12  & -0.16 & 0.09  & -0.23 & 0.28  & 0.01    & 0.00    & -0.01    \\
physiological length                                                        & 0.00  & -0.07 & 0.10  & 0.47  & -0.13 & -0.01    & 0.01    & 0.05      & -0.05 & 0.16  & -0.09 & -0.27 & 0.30  & -0.04    & -0.03    & 0.00    \\
physiological width                                                         & -0.14 & -0.25 & 0.16  & 0.00 & -0.07 & 0.05     & 0.01     & 0.02      & 0.27  & -0.13 & 0.09  & 0.05  & 0.05  & -0.04    & -0.02    & 0.02      \\
aspect ratio                                                                & -0.14 & -0.21 & 0.11  & -0.24 & -0.01 & -31.15     & 0.92      & 5.74        & 0.22  & -0.18 & 0.11  & 0.11  & -0.07 & -0.87     & -12.64      & -14.56       \\
rectangularity                                                              & -0.15 & 0.05  & -0.21 & -0.05 & 0.11  & 39.81       & -9.31      & -54.21       & -0.03 & 0.25  & -0.13 & 0.10  & -0.11 & -14.56      & -1.64      & 33.27        \\
circularity                                                                 & -0.25 & 0.02  & -0.20 & 0.02  & -0.02 & -46.07      & -23.83      & 20.01        & 0.24  & 0.13  & -0.05 & 0.18  & -0.06 & -2.94      & -2.04      & 9.07        \\
compactness                                                                 & 0.23  & -0.06 & 0.23  & -0.01 & 0.03  & 0.14      & 0.48      & -0.61      & -0.22 & -0.07 & 0.03  & -0.18 & -0.02 & 0.20      & -0.09    & 0.05      \\
NF                                                                          & 0.16  & 0.20  & -0.12 & 0.20  & 0.02  & -16.21      & 6.44       & 10.63        & -0.24 & 0.01  & -0.02 & -0.17 & -0.03 & -3.10      & 4.49       & 1.31        \\
Perimeter ratio diameter                                                    & 0.10  & -0.18 & 0.28  & -0.16 & 0.04  & 2.36       & -5.70      & -7.27       & 0.11  & -0.29 & 0.16  & 0.05  & -0.02 & 4.53       & 4.16       & 2.34        \\
Perimeter ratio length                                                      & 0.25  & 0.10  & 0.04  & 0.12  & 0.05  & 3.18       & -2.18      & -3.87       & -0.23 & 0.00  & -0.01 & -0.17 & -0.03 & 1.75       & -1.91      & -0.73      \\
Perimeter ratio lw                                                          & 0.21  & -0.08 & 0.22  & -0.08 & 0.08  & 16.86       & 2.38       & 54.00        & -0.13 & -0.26 & 0.12  & -0.12 & 0.01  & -37.98      & -19.58      & -5.38       \\
No of Convex points                                                         & -0.20 & -0.01 & 0.05  & -0.12 & 0.05  & 0.08     & -0.01    & 0.07      & 0.06  & 0.20  & -0.12 & 0.11  & 0.02  & 0.00    & 0.02     & -0.01    \\
perimeter convexity                                                         & -0.18 & 0.12  & -0.26 & 0.10  & -0.03 & 63.15       & 21.37       & 3.69       & -0.04 & 0.31  & -0.16 & 0.04  & -0.04 & -6.46      & -48.53      & -60.23       \\
area convexity                                                              & 0.23  & -0.08 & 0.19  & 0.02  & 0.04  & -7.31      & -17.58     & 39.50        & -0.04 & -0.31 & -0.31 & -0.09 & 0.03  & -4.75      & -5.18      & -10.42       \\
area ratio convexity                                                        & -0.24 & 0.08  & -0.19 & -0.03 & -0.04 & 23.67       & 12.15       & 74.99        & 0.04  & 0.31  & -0.16 & 0.09  & -0.03 & -16.03      & -30.33      & -40.54       \\
equivalent diameter                                                         & -0.18 & -0.20 & 0.08  & 0.19  & -0.08 & -0.03    & 0.03     & -0.02     & 0.29  & 0.04  & 0.00  & 0.01  & 0.15  & 0.03     & 0.00   & -0.01    \\
cx                                                                          & -0.06 & -0.05 & -0.04 & 0.12  & 0.00 & -0.00   & 0.00   & 0.00     & -0.03 & 0.10  & -0.05 & 0.01  & 0.07  & 0.00   & 0.00   & 0.00    \\
cy                                                                          & -0.16 & -0.17 & 0.05  & -0.12 & 0.03  & 0.01     & 0.01     & -0.01     & 0.01  & 0.05  & 0.00  & -0.04 & 0.09  & 0.00   & 0.00    & 0.00    \\
eccentricity                                                                & 0.09  & 0.20  & -0.17 & 0.24  & -0.01 & -11.57      & -1.28      & -1.46       & -0.15 & 0.22  & -0.13 & -0.10 & 0.07  & 0.02     & 4.63       & 5.11       \\
contrast                                                                    & -0.18 & -0.05 & 0.03  & -0.11 & -0.04 & -0.01    & 0.01    & 0.04      & 0.12  & 0.07  & -0.03 & -0.19 & 0.19  & 0.01    & -0.00   & 0.00    \\
correlation texture                                                         & 0.15  & 0.02  & -0.02 & 0.18  & 0.05  & -0.04    & -74.65      & 653.77        & -0.02 & -0.01 & 0.02  & 0.26  & -0.18 & 162.60       & 70.64       & 89.69       \\
inverse difference moments                                                  & 0.18  & 0.17  & -0.02 & -0.01 & 0.27  & -0.50     & 1.33       & -6.73       & -0.28 & -0.041 & 0.01  & 0.06  & -0.19 & -3.13      & 13.77       & 13.95        \\
entropy                                                                     & -0.17 & -0.18 & -0.01 & 0.12  & -0.26 & -0.25     & -0.07    & -0.14      & 0.28  & 0.05  & -0.01 & -0.06 & 0.18  & 0.14      & 1.26       & 1.41        \\
Mean Red value                                                              & 0.13  & 0.11  & -0.02 & -0.03 & -0.43 & -0.01  & 0.00   & 0.01     & 0.17  & 0.06  & 0.00 & -0.24 & -0.34 & -0.02    & -0.03    & 0.03      \\
Mean Green value                                                             & 0.17  & 0.17  & -0.02 & -0.21 & -0.16 & 0.00  & 0.03     & -0.02     & 0.23  & 0.06  & 0.00 & -0.20 & -0.24 & -0.01   & 0.03     & -0.02     \\
Mean Blue value                                                             & 0.13  & 0.13  & -0.03 & -0.10 & -0.42 & 0.01   & 0.00   & 0.01     & 0.19  & 0.06  & 0.00 & -0.24 & -0.31 & 0.02     & 0.00    & 0.01     \\
\begin{tabular}[c]{@{}l@{}}SD Red value\end{tabular}   & 0.04  & 0.04  & 0.00  & -0.12 & -0.44 & 0.01   & -0.02    & -0.02     & 0.16  & 0.06  & 0.00 & -0.26 & -0.32 & 0.00   & -0.04    & -0.03     \\
\begin{tabular}[c]{@{}l@{}}SD Green value\end{tabular} & -0.15 & -0.03 & 0.04  & 0.19  & 0.24  & 0.02   & -0.05    & -0.02     & 0.19  & -0.03 & 0.02  & 0.06  & 0.20  & 0.00   & -0.01   & 0.01      \\
\begin{tabular}[c]{@{}l@{}}SD  Blue values\end{tabular} & -0.09 & 0.00 & 0.05  & -0.07 & 0.24  & 0.00  & -0.02    & 0.02      & 0.21  & 0.05  & 0.00 & -0.25 & -0.21 & 0.01   & 0.06    & 0.00    \\
correlation                                                                 & 0.00 & 0.01  & 0.00  & -0.05 & 0.08  & 0.16   & -0.13     & 0.58       & 0.03  & 0.10  & 0.03  & 0.23  & 0.02  & -0.04    & -0.23     & 0.21       \\ \hline
                                                                &  &  &   &  &  &    &      &       &  &  & &   &  &    &   &     \\ \hline
Cummulative proportion  - PCA                                                              & 0.24 & 0.43  & 0.54  & 0.61 & 0.66  &    &     &      & 0.21  & 0.36  & 0.49  & 0.58  & 0.64  &    &      &    \\ \hline
\end{tabular}%
}
\caption{Summary of PCA \& LDA coefficients (shape-wise classification)}
\label{tab:pcaldasum}
\end{table}

\begin{figure}
\centering
\includegraphics[width=0.9\columnwidth]{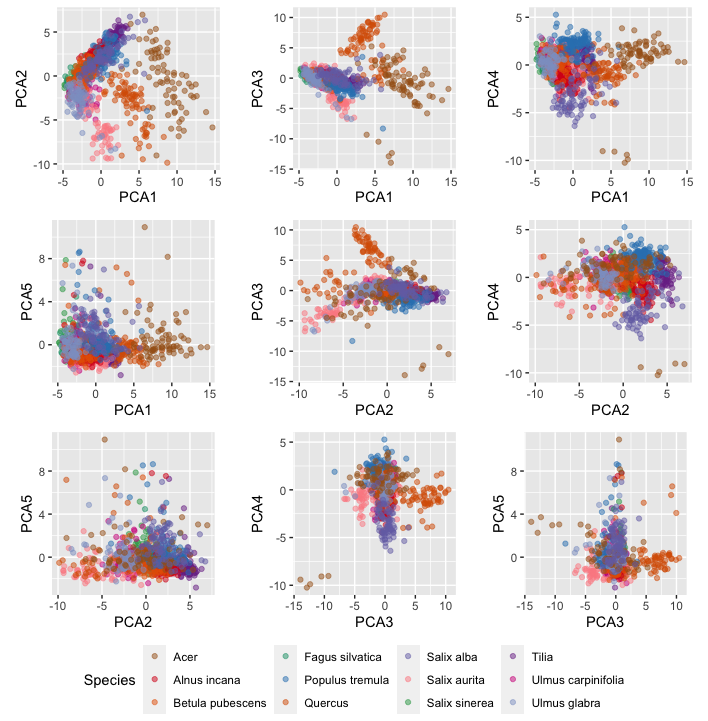}
\caption{\label{pcaswedishsp}Distribution of Swedish leaf images on the
principal component analysis-based projection space. All
projected points are coloured according to their species labels.}
\includegraphics[width=0.8\columnwidth]{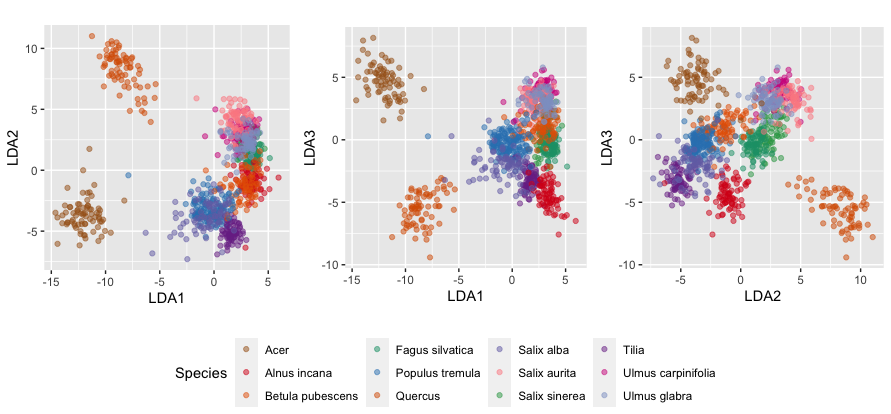}
\caption{\label{ldaswedishsp}Distribution of Swedish leaf images on linear discriminant analysis based projection space. All
projected points are coloured according to their species labels.}
\end{figure}

\begin{figure}
\centering
\includegraphics[width=0.9\columnwidth]{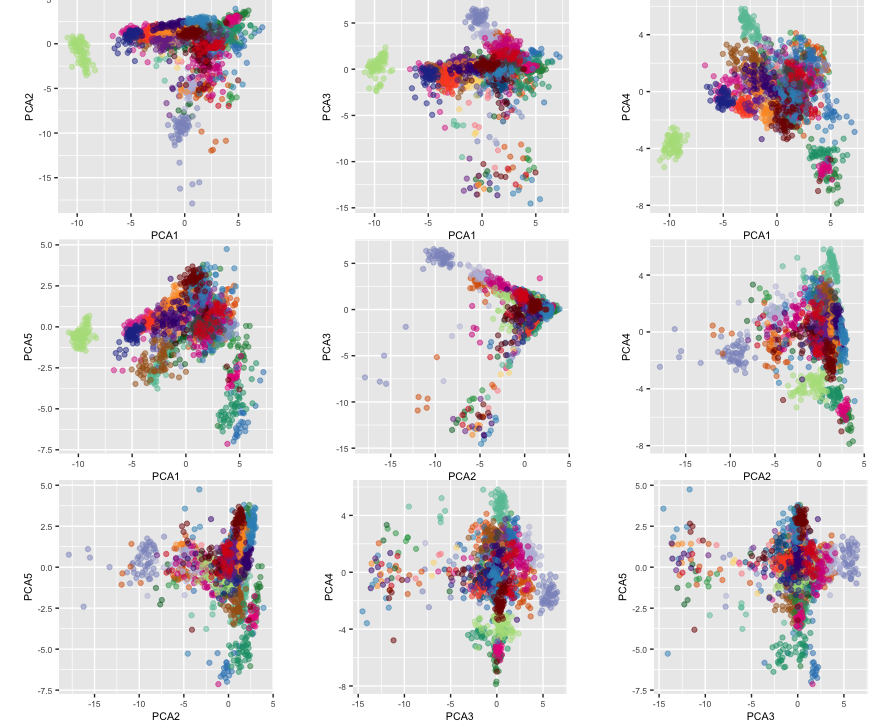}
\caption{\label{pcaflaviasp}Distribution of Flavia leaf images on the
principal component analysis-based projection space. All
projected points are coloured according to their species labels.}
\includegraphics[width=0.8\columnwidth]{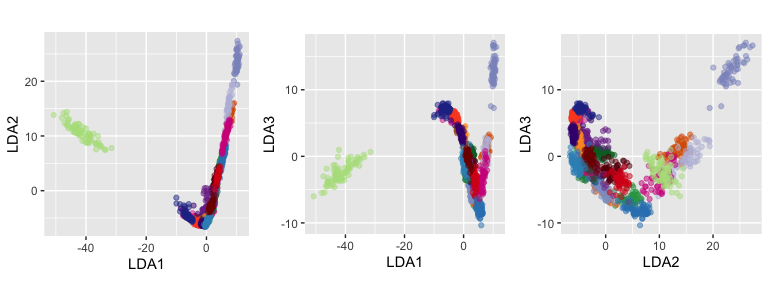}
\includegraphics[width=0.8\columnwidth]{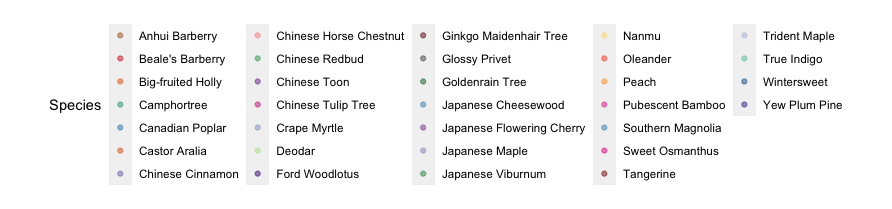}
\caption{\label{ldaflaviasp}Distribution of Flavia leaf images on linear discriminant analysis based projection space. All
projected points are coloured according to their species labels.}
\end{figure}

\newpage

\begin{table}[!ht]
\resizebox{\textwidth}{!}{%
\begin{tabular}{lrrrrrrrrrrrrrrrr}
\hline
\multirow{2}{*}{\begin{tabular}[c]{@{}l@{}}Feature\\ Name\end{tabular}} & \multicolumn{8}{c}{Swedish}                                                                                                                                                                                           & \multicolumn{8}{c}{Flavia}                                                                                                                                                                                            \\ \cline{2-17} 
                                                                        & \multicolumn{1}{l}{PCA1} & \multicolumn{1}{l}{PCA2} & \multicolumn{1}{l}{PCA3} & \multicolumn{1}{l}{PCA4} & \multicolumn{1}{l}{PCA5} & \multicolumn{1}{l}{LDA1} & \multicolumn{1}{l}{LDA2} & \multicolumn{1}{l}{LDA3} & \multicolumn{1}{l}{PCA1} & \multicolumn{1}{l}{PCA2} & \multicolumn{1}{l}{PCA3} & \multicolumn{1}{l}{PCA4} & \multicolumn{1}{l}{PCA5} & \multicolumn{1}{l}{LDA1} & \multicolumn{1}{l}{LDA2} & \multicolumn{1}{l}{LDA3} \\ \hline
Skewed polar                                                            & 0.05                     & -0.08                    & -0.11                    & -0.03                    & 0.03                     & 1.16                     & -0.03                    & -2.37                    & 0.02                     & -0.03                    & -0.10                    & 0.09                     & -0.02                    & 1.02                     & -0.42                    & 1.00                     \\
Clumpy polar                                                            & 0.04                     & -0.07                    & -0.07                    & -0.01                    & -0.03                    & 0.86                     & 0.09                     & 0.20                     & -0.02                    & -0.03                    & -0.06                    & -0.01                    & -0.02                    & 3.05                     & 1.59                     & -1.89                    \\
Sparse polar                                                            & 0.10                     & -0.21                    & -0.16                    & -0.07                    & -0.05                    & -45.51                   & 88.66                    & -115.87                  & 0.02                     & -0.15                    & -0.29                    & 0.02                     & 0.00                     & -31.04                   & -5.91                    & -20.03                   \\
Striated polar                                                          & -0.13                    & 0.22                     & 0.21                     & 0.01                     & -0.02                    & -2.37                    & -0.94                    & -3.69                    & -0.05                    & 0.14                     & 0.31                     & -0.07                    & 0.01                     & 1.88                     & 0.09                     & -0.39                    \\
Convex polar                                                            & 0.12                     & -0.21                    & -0.18                    & -0.06                    & -0.03                    & 19.81                    & -30.88                   & 21.25                    & 0.04                     & -0.16                    & -0.32                    & 0.03                     & 0.03                     & -4.90                    & -1.27                    & -1.68                    \\
Skinny polar                                                            & 0.01                     & -0.09                    & -0.04                    & -0.02                    & 0.08                     & -0.71                    & -0.03                    & -0.45                    & -0.01                    & -0.02                    & -0.08                    & 0.05                     & 0.00                     & -0.02                    & -0.12                    & -0.01                    \\
Stringy polar                                                           & -0.15                    & 0.14                     & 0.19                     & -0.01                    & -0.10                    & 7.21                     & 4.40                     & -1.40                    & -0.05                    & 0.15                     & 0.28                     & -0.07                    & 0.01                     & 2.70                     & 1.67                     & 0.13                     \\
Monotonic polar                                                         & 0.09                     & 0.02                     & -0.05                    & 0.14                     & 0.17                     & -1.09                    & -1.49                    & 0.08                     & 0.07                     & 0.06                     & 0.05                     & -0.09                    & 0.18                     & 0.63                     & 0.51                     & 1.28                     \\
Skewed contour                                                          & 0.07                     & -0.16                    & -0.15                    & 0.01                     & 0.03                     & -0.49                    & 0.21                     & 1.88                     & 0.01                     & -0.04                    & -0.09                    & -0.04                    & -0.06                    & -0.36                    & 1.07                     & 1.10                     \\
Clumpy contour                                                          & 0.04                     & -0.05                    & -0.08                    & 0.05                     & 0.05                     & 1.50                     & 2.49                     & 8.76                     & 0.02                     & -0.02                    & -0.04                    & -0.01                    & 0.00                     & -2.03                    & 0.41                     & 2.33                     \\
Sparse contour                                                          & 0.16                     & -0.18                    & -0.169                   & -0.02                    & 0.01                     & -27.87                   & 0.52                     & 24.01                    & 0.02                     & -0.16                    & -0.29                    & 0.00                     & 0.02                     & 11.62                    & 50.00                    & 30.38                    \\
Striated contour                                                        & -0.15                    & 0.21                     & 0.20                     & 0.00                     & -0.02                    & -0.31                    & 1.36                     & -1.52                    & -0.04                    & 0.15                     & 0.31                     & 0.01                     & 0.02                     & -0.07                    & 2.72                     & 1.60                     \\
Convex contour                                                          & 0.12                     & -0.27                    & -0.17                    & -0.05                    & -0.03                    & -1.93                    & 24.62                    & -44.83                   & 0.04                     & 0.01                     & -0.05                    & 0.09                     & 0.10                     & 0.82                     & -0.12                    & 2.21                     \\
Skinny contour                                                          & 0.04                     & -0.10                    & -0.07                    & 0.04                     & 0.03                     & 0.55                     & 0.57                     & -0.33                    & -0.03                    & 0.00                     & -0.13                    & -0.12                    & 0.04                     & 0.23                     & 0.35                     & -0.34                    \\
Stringy contour                                                         & -0.13                    & 0.08                     & 0.13                     & -0.02                    & -0.09                    & 0.83                     & -0.75                    & 0.52                     & -0.04                    & 0.15                     & 0.30                     & -0.01                    & -0.03                    & -2.74                    & -4.56                    & -2.41                    \\
Monotonic contour                                                       & 0.01                     & 0.04                     & 0.01                     & -0.17                    & 0.07                     & -0.96                    & 0.15                     & 0.56                     & 0.05                     & 0.03                     & 0.08                     & 0.25                     & 0.05                     & -0.04                    & -0.16                    & -1.62                    \\
No of max points                                                        & 0.08                     & -0.21                    & -0.20                    & 0.00                     & -0.04                    & 0.02                     & 0.05                     & 0.05                     & -0.02                    & -0.15                    & -0.23                    & -0.07                    & -0.03                    & -0.02                    & -0.06                    & 0.00                     \\
No of min points                                                        & 0.07                     & -0.21                    & -0.18                    & -0.01                    & -0.06                    & -0.01                    & 0.00                     & 0.02                     & -0.02                    & -0.16                    & -0.22                    & -0.12                    & -0.03                    & 0.01                     & 0.05                     & -0.05                    \\
diameter                                                                & 0.01                     & -0.08                    & 0.11                     & 0.45                     & -0.12                    & -0.04                    & -0.02                    & -0.02                    & -0.03                    & 0.15                     & -0.09                    & -0.29                    & 0.29                     & 0.03                     & 0.00                     & -0.01                    \\
area                                                                    & -0.19                    & -0.21                    & 0.05                     & 0.16                     & -0.07                    & 0.00                     & 0.00                     & 0.00                     & 0.28                     & 0.04                     & 0.00                     & -0.01                    & 0.15                     & 0.00                     & 0.00                     & 0.00                     \\
perimeter                                                               & 0.09                     & -0.20                    & 0.29                     & 0.11                     & -0.03                    & 0.00                     & 0.00                     & 0.00                     & 0.12                     & -0.16                    & 0.09                     & -0.23                    & 0.28                     & 0.00                     & 0.01                     & 0.00                     \\
physiological length                                                    & 0.00                     & -0.07                    & 0.10                     & 0.47                     & -0.13                    & 0.01                     & 0.02                     & -0.02                    & -0.05                    & 0.16                     & -0.09                    & -0.27                    & 0.30                     & -0.03                    & -0.04                    & 1.22                     \\
physiological width                                                     & -0.14                    & -0.25                    & 0.16                     & 0.00                     & -0.07                    & -0.03                    & 0.05                     & -0.02                    & 0.27                     & -0.13                    & 0.09                     & 0.05                     & 0.05                     & -0.03                    & -0.04                    & -3.57                    \\
aspect ratio                                                            & -0.14                    & -0.21                    & 0.12                     & -0.24                    & -0.01                    & 17.34                    & -11.17                   & 14.83                    & 0.22                     & -0.18                    & 0.11                     & 0.12                     & -0.07                    & -10.67                   & 8.97                     & 9.08                     \\
rectangularity                                                          & -0.15                    & 0.05                     & -0.21                    & -0.05                    & 0.11                     & 1.17                     & 3.71                     & -29.06                   & -0.03                    & 0.25                     & -0.13                    & 0.10                     & -0.12                    & -42.29                   & -3.09                    & -32.86                   \\
circularity                                                             & -0.25                    & 0.02                     & -0.20                    & 0.02                     & -0.02                    & -32.21                   & 38.61                    & 21.19                    & 0.24                     & 0.13                     & -0.05                    & 0.18                     & -0.06                    & 39.73                    & -49.27                   & -0.73                    \\
compactness                                                             & 0.23                     & -0.06                    & 0.23                     & -0.01                    & 0.03                     & 82.91                    & -0.24                    & -0.12                    & -0.22                    & -0.07                    & 0.03                     & -0.18                    & -0.02                    & -0.09                    & 0.05                     & -0.07                    \\
NF                                                                      & 0.16                     & 0.20                     & -0.12                    & 0.20                     & 0.02                     & 1.21                     & 19.88                    & -1.71                    & -0.24                    & 0.01                     & -0.02                    & -0.17                    & -0.03                    & 3.10                     & -2.01                    & 2.17                     \\
Perimeter ratio diameter                                                & 0.10                     & -0.18                    & 0.28                     & -0.16                    & 0.04                     & -10.11                   & 6.76                     & -6.66                    & 0.12                     & -0.29                    & 0.16                     & 0.05                     & -0.02                    & 6.48                     & -0.09                    & 0.48                     \\
Perimeter ratio length                                                  & 0.25                     & 0.10                     & 0.04                     & 0.12                     & 0.05                     & -1.97                    & -4.69                    & 5.32                     & -0.23                    & 0.00                     & -0.01                    & -0.17                    & -0.03                    & -0.91                    & 1.03                     & -0.74                    \\
Perimeter ratio lw                                                      & 0.21                     & -0.08                    & 0.22                     & -0.08                    & 0.08                     & -19.35                   & -0.07                    & -1.01                    & -0.13                    & -0.26                    & 0.12                     & -0.12                    & 0.01                     & 2.40                     & -48.35                   & 19.77                    \\
No of Convex points                                                     & -0.20                    & -0.01                    & 0.05                     & -0.12                    & 0.05                     & 0.01                     & 0.02                     & -0.04                    & 0.06                     & 0.20                     & 0.12                     & 0.11                     & 0.02                     & -0.03                    & -0.03                    & 0.06                     \\
perimeter convexity                                                     & -0.18                    & 0.12                     & -0.26                    & 0.10                     & -0.03                    & 30.96                    & -89.46                   & -25.06                   & -0.04                    & 0.31                     & -0.16                    & 0.05                     & -0.04                    & -55.44                   & -10.72                   & 28.83                    \\
area convexity                                                          & 0.23                     & -0.08                    & 0.19                     & 0.02                     & 0.04                     & 32.73                    & -15.20                   & 24.97                    & -0.04                    & -0.31                    & 0.16                     & -0.09                    & 0.03                     & 16.76                    & -4.85                    & 20.29                    \\
area ratio convexity                                                    & -0.24                    & 0.078                    & -0.19                    & -0.03                    & -0.04                    & 95.81                    & -29.35                   & 1.23                     & 0.04                     & 0.31                     & -0.16                    & 0.09                     & -0.03                    & 14.55                    & -1.34                    & 41.66                    \\
equivalent diameter                                                     & -0.18                    & -0.20                    & 0.081                    & 0.19                     & -0.08                    & 0.09                     & -0.04                    & 0.05                     & 0.29                     & 0.04                     & 0.00                     & 0.01                     & 0.15                     & 0.02                     & 0.08                     & 0.00                     \\
cx                                                                      & -0.06                    & -0.05                    & -0.04                    & 0.12                     & -0.01                    & 0.00                     & 0.00                     & 0.00                     & -0.03                    & 0.10                     & -0.05                    & 0.01                     & 0.07                     & 0.00                     & 0.00                     & 0.00                     \\
cy                                                                      & -0.16                    & -0.17                    & 0.05                     & -0.12                    & 0.03                     & 0.00                     & 0.00                     & 0.01                     & 0.10                     & 0.05                     & 0.00                     & -0.04                    & 0.09                     & 0.00                     & 0.00                     & 0.00                     \\
eccentriciry                                                            & 0.09                     & 0.20                     & -0.17                    & 0.24                     & -0.01                    & 1.26                     & 3.10                     & 5.13                     & -0.15                    & 0.22                     & -0.13                    & -0.10                    & 0.07                     & 5.37                     & 1.57                     & -2.01                    \\
contrast                                                                & -0.18                    & -0.05                    & 0.03                     & -0.11                    & -0.04                    & -0.09                    & -0.03                    & 0.05                     & 0.12                     & 0.07                     & -0.03                    & -0.19                    & 0.19                     & 0.01                     & 0.02                     & 0.04                     \\
correlation texture                                                     & 0.15                     & 0.02                     & -0.02                    & 0.18                     & 0.05                     & -185.14                  & -233.28                  & 143.05                   & -0.02                    & -0.01                    & 0.02                     & 0.26                     & -0.18                    & 50.17                    & 375.80                   & 296.79                   \\
inverse difference moments                                              & 0.18                     & 0.17                     & -0.02                    & -0.01                    & 0.27                     & -17.60                   & -1.87                    & 31.96                    & -0.28                    & -0.04                    & 0.01                     & 0.06                     & -0.19                    & -7.20                    & -1.70                    & 39.29                    \\
entropy                                                                 & -0.17                    & -0.18                    & -0.01                    & 0.12                     & -0.26                    & 0.01                     & 0.07                     & 2.32                     & 0.28                     & 0.05                     & -0.01                    & -0.06                    & 0.18                     & -0.36                    & 2.47                     & 4.72                     \\
Mean red value                                                          & 0.13                     & 0.11                     & -0.02                    & -0.03                    & -0.43                    & -0.01                    & 0.00                     & 0.00                     & 0.17                     & 0.06                     & 0.00                     & -0.24                    & -0.34                    & 0.02                     & -0.04                    & 0.11                     \\
Mean green value                                                        & 0.17                     & 0.17                     & -0.02                    & -0.21                    & -0.16                    & -0.02                    & 0.00                     & 0.03                     & 0.23                     & 0.06                     & 0.00                     & -0.20                    & -0.24                    & 0.03                     & 0.00                     & 0.04                     \\
Mean blue value                                                         & 0.13                     & 0.13                     & -0.03                    & -0.10                    & -0.42                    & -0.01                    & -0.01                    & -0.01                    & 0.19                     & 0.06                     & 0.00                     & -0.24                    & -0.31                    & -0.04                    & 0.04                     & -0.14                    \\
SD red value                                                            & 0.04                     & 0.04                     & 0.00                     & -0.12                    & -0.44                    & 0.02                     & 0.00                     & 0.00                     & 0.16                     & 0.06                     & 0.00                     & -0.26                    & -0.32                    & -0.03                    & -0.05                    & -0.12                    \\
SD blue value                                                           & -0.15                    & -0.03                    & 0.04                     & 0.19                     & 0.24                     & 0.05                     & -0.02                    & -0.06                    & 0.19                     & -0.03                    & 0.02                     & 0.06                     & 0.20                     & 0.00                     & -0.04                    & -0.05                    \\
SD green value                                                          & -0.09                    & 0.00                     & 0.05                     & -0.07                    & 0.24                     & -0.01                    & 0.01                     & -0.02                    & 0.21                     & 0.05                     & 0.00                     & -0.25                    & -0.21                    & 0.01                     & 0.07                     & 0.12                     \\
correlation                                                             & 0.00                     & 0.01                     & 0.00                     & -0.05                    & 0.08                     & -0.28                    & 0.26                     & -0.41                    & 0.03                     & 0.10                     & 0.03                     & 0.23                     & 0.02                     & 0.04                     & -0.12                    & 0.23                     \\ \hline
                                                                &  &  &   &  &  &    &      &       &  &  & &   &  &    &   &     \\ \hline
Cummulative proportion  - PCA                                                              & 0.24 & 0.43  & 0.54  & 0.61 & 0.66  &    &     &      & 0.21  & 0.36  & 0.49  & 0.58  & 0.64  &    &      &    \\ \hline
\end{tabular}%
}
\caption{Summary of PCA \& LDA coefficients (species-wise classification)}
\label{tab:pcaldasumsp}
\end{table}

\hypertarget{swedish-dataset}{%
\subsubsection{Swedish Dataset}\label{swedish-dataset}}

The visualization of PCA projections of the Swedish dataset is shown
in Figure \ref{pcaswedish}. The first three principal components (PCs)
accounting for approximately 43\% of the total variance in the original
data, while the first 5 PCs account for 66\% of the total variance in
the original data. Hence, the first five PCs are plotted against each
other to visualize data in the PCA space. The LDA projections of Swedish
data are shown in Figure \ref{ldaswedish}. According to the LDA results
on the Swedish dataset, LDA1, LDA2, and LDA3 show a clear separation of
shapes of leaf images.


\hypertarget{flavia-dataset}{%
\subsubsection{Flavia Dataset}\label{flavia-dataset}}

~~~~~The LDA projections of Flavia data are shown in Figure
\ref{ldaflavia}. The visualization of PCA projections of the Flavia dataset
is shown in Figure \ref{pcaflavia}. The first three principal components
(PCs) accounting for approximately 49\% of the total variance in the
original data, while the first 5 PCs account for 64\% of the total
variance in the original data. Hence, the first five PCs are plotted
against each other to visualize data in the PCA projection space.
According to the LDA results on the Flavia dataset, LDA1, LDA2, and LDA3
shows a clear separation of shapes of leaf images. Under both
experimental settings, class separation is more clearly on the LDA space
than the PCA space. The reason could be LDA is a supervised learning
algorithm while PCA space is an unsupervised learning algorithm.

According to both PCA and LDA visualizations on Swedish and Flavia data
sets, we can see a clear separation of classes in their corresponding
projection spaces. This reveals our features are capable of
distinguishing classes under both supervised learning and unsupervised
learning settings. The class separations are clearly visible under both small numbers of class labels and a large number of class labels. The PCA and LDA coefficients for shape-wise classification and species-wise classifications are shown in Table \ref{tab:pcaldasum} and Table \ref{tab:pcaldasumsp} respectively.

\hypertarget{discussion-and-conclusions}{%
\section{Discussion and Conclusions}\label{discussion-and-conclusions}}

~~~~~~In this paper we introduce computer-aided, interpretable features
for image recognition. There are four main categories of features that
are used to classify leaf images. Many research was based on shape,
color, and texture-based image features. In this research paper, we introduce a new feature category called scagnostics for image classification. Other than
that correlation of cartesian coordinate, number of convex points,
number of minimum and maximum points are introduced as new shape
features. We explore the ability of features to discriminate the classes
of interest under supervised learning and unsupervised learning settings
using principal component analysis and linear discriminant analysis.
Under both experimental settings, a clear separation of classes is visible
in their projection spaces. In our next paper, we introduce a
meta-learning algorithm to classify plant species using the features
introduced in this paper. A more detailed look at the most important
features and approach of classification plant species will be presented
in our next paper. We hope our analysis opens various research agendas
in the field of image recognition. Reproducible research code to
reproduce the results of this paper and codes to compute features are
available here: \url{https://github.com/SMART-Research/leaffeatures_paper}.

\newpage
\bibliographystyle{unsrtnat}
\nocite{*}
\bibliography{references}  

\begin{thebibliography}{25}
\providecommand{\natexlab}[1]{#1}
\providecommand{\url}[1]{\texttt{#1}}
\expandafter\ifx\csname urlstyle\endcsname\relax
  \providecommand{\doi}[1]{doi: #1}\else
  \providecommand{\doi}{doi: \begingroup \urlstyle{rm}\Url}\fi

\bibitem[W{\"a}ldchen and M{\"a}der(2018)]{articlee}
Jana W{\"a}ldchen and Patrick M{\"a}der.
\newblock Plant species identification using computer vision techniques: A
  systematic literature review.
\newblock \emph{Archives of Computational Methods in Engineering}, 25\penalty0
  (2):\penalty0 507--543, 2018.

\bibitem[Anantrasirichai et~al.(2017)Anantrasirichai, Hannuna, and
  Canagarajah]{DBLP}
Nantheera Anantrasirichai, Sion Hannuna, and Nishan Canagarajah.
\newblock Automatic leaf extraction from outdoor images.
\newblock \emph{arXiv preprint arXiv:1709.06437}, 2017.

\bibitem[Wu et~al.(2007{\natexlab{a}})Wu, Bao, Xu, Wang, Chang, and
  Xiang]{4458016}
Stephen~Gang Wu, Forrest~Sheng Bao, Eric~You Xu, Yu-Xuan Wang, Yi-Fan Chang,
  and Qiao-Liang Xiang.
\newblock A leaf recognition algorithm for plant classification using
  probabilistic neural network.
\newblock In \emph{2007 IEEE international symposium on signal processing and
  information technology}, pages 11--16. IEEE, 2007{\natexlab{a}}.

\bibitem[Azlah et~al.(2019)Azlah, Chua, Rahmad, Abdullah, and
  Wan~Alwi]{articlepl}
Muhammad Azfar~Firdaus Azlah, Lee~Suan Chua, Fakhrul~Razan Rahmad, Farah~Izana
  Abdullah, and Sharifah~Rafidah Wan~Alwi.
\newblock Review on techniques for plant leaf classification and recognition.
\newblock \emph{Computers}, 8\penalty0 (4):\penalty0 77, 2019.

\bibitem[Herdiyeni and Wahyuni(2012)]{inproceedings}
Yeni Herdiyeni and Ni~Kadek~Sri Wahyuni.
\newblock Mobile application for indonesian medicinal plants identification
  using fuzzy local binary pattern and fuzzy color histogram.
\newblock In \emph{2012 International Conference on Advanced Computer Science
  and Information Systems (ICACSIS)}, pages 301--306. IEEE, 2012.

\bibitem[Gonzalez and Woods(2006)]{book1}
Rafael~C. Gonzalez and Richard~E. Woods.
\newblock \emph{Digital image processing}.
\newblock Prentice-Hall, Inc., 2006.

\bibitem[{Goyal} et~al.(2018){Goyal}, {Kapil}, and {Kumar}]{8675114}
N.~{Goyal}, {Kapil}, and N.~{Kumar}.
\newblock Plant species identification using leaf image retrieval: A study.
\newblock In \emph{2018 International Conference on Computing, Power and
  Communication Technologies (GUCON)}, pages 405--411, 2018.

\bibitem[Wu et~al.(2007{\natexlab{b}})Wu, Bao, Xu, Wang, Chang, and
  Xiang]{wu2007leaf}
Stephen~Gang Wu, Forrest~Sheng Bao, Eric~You Xu, Yu-Xuan Wang, Yi-Fan Chang,
  and Qiao-Liang Xiang.
\newblock A leaf recognition algorithm for plant classification using
  probabilistic neural network.
\newblock In \emph{2007 {IEEE} International Symposium on Signal Processing and
  Information Technology}, pages 11--16. IEEE, 2007{\natexlab{b}}.

\bibitem[S{\"o}derkvist(2001)]{soderkvist2001computer}
Oskar S{\"o}derkvist.
\newblock Computer vision classification of leaves from swedish trees, 2001.

\bibitem[Mingqiang et~al.(2008)Mingqiang, Kidiyo, and Joseph]{article7}
Yang Mingqiang, Kpalma Kidiyo, and Ronsin Joseph.
\newblock A survey of shape feature extraction techniques.
\newblock \emph{Pattern Recognition}, 15\penalty0 (7):\penalty0 43--90, 2008.

\bibitem[Sun et~al.(2017)Sun, Liu, Wang, and Zhang]{sun2017deep}
Yu~Sun, Yuan Liu, Guan Wang, and Haiyan Zhang.
\newblock Deep learning for plant identification in natural environment.
\newblock \emph{Computational intelligence and neuroscience}, 2017, 2017.

\bibitem[Wilkinson et~al.(2005)Wilkinson, Anand, and Grossman]{inproceedings44}
Leland Wilkinson, Anushka Anand, and Robert Grossman.
\newblock Graph-theoretic scagnostics.
\newblock In \emph{IEEE Symposium on Information Visualization (InfoVis 05)},
  pages 157--158. IEEE Computer Society, 2005.

\bibitem[Dean(1999)]{article31}
C~Dean.
\newblock \emph{Quantitative description and automated classification of
  cellular protein localization patterns in fluorescence microscope images of
  mammalian cells}.
\newblock PhD thesis, PhD thesis, Carnegie Mellon University, 1999.

\bibitem[Caglayan et~al.(2013)Caglayan, Guclu, and Can]{inproceedings1}
Ali Caglayan, Oguzhan Guclu, and Ahmet~Burak Can.
\newblock A plant recognition approach using shape and color features in leaf
  images.
\newblock In \emph{International Conference on Image Analysis and Processing},
  pages 161--170. Springer, 2013.

\bibitem[Kodituwakku and Selvarajah(2004)]{colarticle1}
Saluka~Ranasinghe Kodituwakku and S~Selvarajah.
\newblock Comparison of color features for image retrieval.
\newblock \emph{Indian Journal of Computer Science and Engineering}, 1\penalty0
  (3):\penalty0 207--211, 2004.

\bibitem[Wilkinson and Wills(2008)]{article37}
Leland Wilkinson and Graham Wills.
\newblock Scagnostics distribution.
\newblock \emph{Journal of Computational and Graphical Statistics},
  17:\penalty0 473--491, 06 2008.
\newblock \doi{10.1198/106186008X320465}.

\bibitem[Hartigan and Mohanty(1992)]{hartigan1992runt}
John~A Hartigan and Surya Mohanty.
\newblock The runt test for multimodality.
\newblock \emph{Journal of Classification}, 9\penalty0 (1):\penalty0 63--70,
  1992.

\bibitem[Lakshika and Talagala(2021)]{medlea}
Jayani P.~G. Lakshika and Thiyanga~S. Talagala.
\newblock Medlea: Morphological and structural features of medicinal leaves.
\newblock 2021.
\newblock URL \url{https://CRAN.R-project.org/package=MedLEA}.
\newblock R package version 1.0.1.

\bibitem[Aakif and Khan(2015)]{article10}
Aimen Aakif and Muhammad~Faisal Khan.
\newblock Automatic classification of plants based on their leaves.
\newblock \emph{Biosystems Engineering}, 139:\penalty0 66--75, 2015.

\bibitem[Jeon and Rhee(2017)]{jarticle7}
Wang-Su Jeon and Sang-Yong Rhee.
\newblock Plant leaf recognition using a convolution neural network.
\newblock \emph{International Journal of Fuzzy Logic and Intelligent Systems},
  17\penalty0 (1):\penalty0 26--34, 2017.

\bibitem[Dang and Wilkinson(2014)]{inproceedings33}
Tommy Dang and Leland Wilkinson.
\newblock Scagexplorer: Exploring scatterplots by their scagnostics.
\newblock pages 73--80, 03 2014.
\newblock \doi{10.1109/PacificVis.2014.42}.

\bibitem[Dang et~al.(2012)Dang, Anand, and Wilkinson]{article101}
Tuan~Nhon Dang, Anushka Anand, and Leland Wilkinson.
\newblock Timeseer: Scagnostics for high-dimensional time series.
\newblock \emph{IEEE Transactions on Visualization and Computer Graphics},
  19\penalty0 (3):\penalty0 470--483, 2012.

\bibitem[Kalyoncu and Toygar(2015)]{article5}
Cem Kalyoncu and {\"O}nsen Toygar.
\newblock Geometric leaf classification.
\newblock \emph{Computer Vision and Image Understanding}, 133:\penalty0
  102--109, 2015.

\bibitem[Haralick et~al.(1973)Haralick, Shanmugam, and Dinstein]{articletx}
Robert~M Haralick, Karthikeyan Shanmugam, and Its'~Hak Dinstein.
\newblock Textural features for image classification.
\newblock \emph{IEEE Transactions on systems, man, and cybernetics}, \penalty0
  (6):\penalty0 610--621, 1973.

\bibitem[Bangare et~al.(2015)Bangare, Dubal, Bangare, and
  Patil]{bangare2015reviewing}
Sunil~L Bangare, Amruta Dubal, Pallavi~S Bangare, and ST~Patil.
\newblock Reviewing otsu’s method for image thresholding.
\newblock \emph{International Journal of Applied Engineering Research},
  10\penalty0 (9):\penalty0 21777--21783, 2015.

\end{thebibliography}






\end{document}